\documentclass[preprint,12pt]{elsarticle}

\usepackage{amssymb}
\usepackage{bigstrut}
\usepackage{graphicx}
\graphicspath{{}}
\DeclareGraphicsExtensions{.pdf,.jpeg,.png,.eps}
\usepackage{subfigure} 
\usepackage{algorithm}
\usepackage{algorithmic}
\usepackage[cmex10]{amsmath}
\usepackage{amsmath}
\usepackage{stmaryrd}
\usepackage{multirow}
\usepackage{url}
\usepackage[latin1]{inputenc}
\usepackage{rotating}
\usepackage[svgnames]{xcolor}

\makeatletter
\def\ps@pprintTitle{%
  \let\@oddhead\@empty
  \let\@evenhead\@empty
  \def\@oddfoot{\reset@font\hfil\thepage\hfil}
  \let\@evenfoot\@oddfoot
}
\makeatother

\journal{Neurocomputing}

\begin{document}

\begin{frontmatter}



\title{Double-Base Asymmetric AdaBoost}

\author[uvigo]{Iago Landesa-V\'azquez}
\ead{iagolv@gts.uvigo.es}

\author[uvigo] {Jos\'e Luis Alba-Castro}
\ead{jalba@gts.uvigo.es}

\address[uvigo]{Signal Theory and Communications Department, University of Vigo, Maxwell Street, 36310, Vigo, Spain}
\address{}

\begin{abstract}
Based on the use of different exponential bases to define class-dependent error bounds, a new and highly efficient asymmetric boosting scheme, coined as AdaBoostDB (Double-Base), is proposed. Supported by a fully theoretical derivation procedure, unlike most of the other approaches in the literature, our algorithm preserves all the formal guarantees and properties of original (cost-insensitive) AdaBoost, similarly to the state-of-the-art Cost-Sensitive AdaBoost algorithm. However, the key advantage of AdaBoostDB is that our novel derivation scheme enables an extremely efficient conditional search procedure, dramatically improving and simplifying the training phase of the algorithm. Experiments, both over synthetic and real datasets, reveal that AdaBoostDB is able to save over 99\% training time with regard to Cost-Sensitive AdaBoost, providing the same cost-sensitive results. This computational advantage of AdaBoostDB can make a difference in problems managing huge pools of weak classifiers in which boosting techniques are commonly used. 
\end{abstract}


\end{frontmatter}

\section{Introduction}
\label{sec:Intro}
Boosting algorithms \cite{Schapire90}, with AdaBoost \cite{FreundSchapire97} as epitome, have been an active focus of research since its first publication in the 1990s. Its strong theoretical guarantees together with promising practical results, including robustness against overfitting and ease of implementation, have drawn the attention towards this family of algorithms over the last decade, \cite{Schapire98,SchapireSinger99,Opitz99,Friedman00,MeaseWyner08a,ViolaJones04,GomezVerdejo10} both from the theoretical and practical perspectives.

A plethora of different applications (medical diagnosis, fraud detection, biometrics, disaster prediction\ldots) have implicit classification tasks with well-defined costs depending on the different kinds of mistakes in each possible decision (false positives and false negatives). On the other hand, many problems have imbalanced class priors, so one class is extremely more frequent or easier to sample than the other one. To deal with such scenarios \cite{Elkan01,Weiss03}, classifiers must be capable of focusing their attention in the rare class, instead of searching hypothesis that, trying to fit well to data, end up being driven by the prevalent class.

Several modifications of AdaBoost have been proposed in the literature to deal with \emph{asymmetry} \cite{KarakoulasShawe98,Fan99,Ting00,ViolaJones04,ViolaJones02,Sun07,MasnadiVasconcelos11}. In the well-known Viola and Jones face detector framework, a validation set is used to modify the AdaBoost strong classifier threshold \emph{a posteriori}, in order to adjust false positive and detection rates balance. Nevertheless, as the authors stated, it is not clear if the selected weak classifiers are optimal for the asymmetric goal \cite{ViolaJones02} nor if these modifications preserve AdaBoost training and generalization errors original guarantees \cite{ViolaJones04}. The vast majority of other proposed methods \cite{KarakoulasShawe98,Fan99,Ting00,ViolaJones02,Sun07} try to cope with asymmetry through direct manipulations of the weight updating rule. These proposals, not being a full reformulation of the algorithm for asymmetric scenarios, have been analyzed \cite{MasnadiVasconcelos07,MasnadiVasconcelos11} to be heuristic modifications of AdaBoost. However, two recent contributions have been proposed to deal with the asymmetric boosting problem in a fully theoretical way: On the one hand, the Cost-Sensitive Boosting framework by H. Masnadi-Shirazi and N. Vasconcelos \cite{MasnadiVasconcelos11} drives to an algorithm far more complex and computationally demanding that the original (symmetric) AdaBoost but with strong theoretical guarantees. And on the other hand, the class-conditional description of AdaBoost by I. Landesa-Vázquez and J.L. Alba-Castro \cite{LandesaAlba12}, demonstrates that asymmetric weight initialization is also an effective and theoretically sound way to reach boosted cost-sensitive classifiers. These two theoretical alternatives follow different ``asymmetrizing'' perspectives and drive to different solutions.

In this work we will follow an approach closer to the Cost-Sensitive Boosting framework \cite{MasnadiVasconcelos11}. Though sharing equivalent theoretical roots and guarantees with Cost-Sensitive AdaBoost \cite{MasnadiVasconcelos11}, our proposal entails a new and self-contained analytical framework leading to a novel asymmetric boosting algorithm which we call AdaBoostDB (from AdaBoost with Double-Base). Our approach is based on three distinctive premises: its derivation is inspired by the generalized boosting framework \cite{FreundSchapire99} (unlike the Statistical View of Boosting followed by \cite{MasnadiVasconcelos11}), its error bound is modeled in terms of class-conditional (double) exponential bases, and two parallel class-conditional weight \emph{subdistributions} are used and updated during the boosting iterations. As a result, from a different (thought theoretically equivalent) perspective, and following a completely different derivation path, we reach an algorithm able to find the same solution as Cost-Sensitive AdaBoost, but in a much more efficient way. Indeed, our approach gives rise to a more tractable mathematical model and enables a searching scheme that dramatically reduces the number of weak classifiers to be evaluated in each iteration.

The paper is organized as follows: In the next section we describe AdaBoost original algorithm and the way asymmetric variations have been proposed in the literature, paying special attention to the Cost-Sensitive AdaBoost algorithm \cite{MasnadiVasconcelos11}. In Section~\ref{sec:AdaBoostDB} AdaBoostDB and the Conditional Search method are derived, explained and discussed. In Section~\ref{sec:Experiments} all the empirical framework and experiments are shown. Finally, Section~\ref{sec:Conclusions} includes the main ideas, conclusions and future research lines drawn from our work.

\section{AdaBoost and Cost}
\label{sec:AdaBoost_Cost}

\subsection{AdaBoost}
\label{subsec:AdaBoost}

Given a space of feature vectors $\mathbf{X}$ and two possible class labels $y\in\{+1,-1\}$, AdaBoost goal is to learn a strong classifier $H(\mathbf{x})$ as a weighted ensemble of weak classifiers $h_{t}(\mathbf{x})$ predicting the label of any instance $\mathbf{x} \in \mathbf{X}$.


\begin{equation}
\label{strong_clas_eqn}
H(x)=
\mathrm{sign}\left(f(\mathbf{x})\right)=
\mathrm{sign}\left(\sum_{t=1}^T\alpha_{t}h_{t}(\mathbf{x})\right)
\end{equation}

From a training set of $n$ examples $\mathbf{x}_i$, each of them labeled as positive $(y_i=1)$ or negative $(y_i=-1)$, and a weight distribution $D_{t}(i)$ defined over them for each learning round $t$, the weak learner must select the best classifier $h_{t}(x)$ according to the labels and weights. Once a weak classifier is selected, it is added to the ensemble modulated by a goodness parameter $\alpha_{t}$ (\ref{alphat_eqn}),  correspondingly updating the weight distribution. Weak hypothesis search is guided to maximize goodness $\alpha_{t}$, which is equivalent to maximize weighted correlation between labels $(y_{i})$ and predictions $(h_{t})$. This procedure can be repeated in an iterative way until a predefined number $T$ of training rounds have been completed or some performance goal is reached:

\begin{equation}
\label{alphat_eqn}
\alpha_{t}=
\frac{1}{2} \ln \left(\frac{1+\sum_{i=1}^{n}D_{t}(i) y_{i} h_{t} (x_{i})}{1-\sum_{i=1}^{n}D_{t}(i) y_{i} h_{t} (x_{i})}\right)
\end{equation}

\begin{equation}
\label{weight_rule_eqn}
\begin{split}
D_{t+1}(i)&= \frac{D_{t}(i)\exp\left(-\alpha_{t} y_{i} h_{t}(x_{i})\right)}{\sum_{i=1}^{n} D_{t}(i) \exp\left(-\alpha_{t} y_{i} h_{t}(x_{i})\right)}\\
&= \frac{D_{t}(i)\exp\left(-\alpha_{t} y_{i} h_{t}(x_{i})\right)}{Z_{t}}
\end{split}
\end{equation}

This scheme can be derived \cite{SchapireSinger99} as a round-by-round (additive) minimization of an exponential bound on the strong classifier training error, coming from the next inequality:

\begin{equation}
\label{bound_ineq_eqn}
H(x_{i})\neq y_{i} \:\Rightarrow\: y_{i} f(x_{i}) \leq 0 \:\Rightarrow\: e^{-y_{i} f(x_{i})} \geq 1
\end{equation}

From now on, as in many other studies, we will focus on the discrete version of AdaBoost for a simpler and more intuitive analysis (which does not prevent our derivations from being also applied to other variations of the algorithm). In this case weak hypothesis are binary $y_{i}\in\{-1,+1\}$, so parameter $\alpha_{t}$ can be rewritten (\ref{alphat2_eqn}) in terms of the weighted error $\varepsilon_{t}$ (\ref{round_err_eqn}) and the weak hypothesis is equivalent to finding the classifier with smaller $\varepsilon_{t}$.

\begin{equation}
\label{alphat2_eqn}
\alpha_{t}=
\frac{1}{2} \ln \left( \frac{1-\varepsilon_{t}}{\varepsilon_{t}}\right)
\end{equation}

\begin{equation}
\label{round_err_eqn}
\varepsilon_{t}=
\sum_{i=1}^{n} D_{t}(i) \llbracket h(x_{i}) \neq y_{i}\rrbracket=
\sum_{\textrm{nok}}D_{t}(i)
\end{equation}

As can be seen, we will follow notation from \cite{SchapireSinger99}, where operator $\llbracket a \rrbracket$ is $1$ when $a$ is true and $0$ otherwise. In addition, for the sake of simplicity, we will use the term `ok' to refer to those training examples in which the result of the weak classifier is right $\{i:h(x_{i})=y_{i}\}$ and `nok' when it is wrong $\{i:h(x_{i})\neq y_{i}\}$. 

\subsection{Cost-Sensitive AdaBoost}
\label{subsec:CSAdaBoost}

As was initially defined, the exponential error bound does not have any direct class-dependent behavior, so several modifications of AdaBoost have been proposed in the literature to enhance this seemingly symmetric nature. Most of the proposed variations \cite{KarakoulasShawe98,Fan99,Ting00,Sun07,ViolaJones02} are based on directly modifying the weight update rule in an asymmetric (class-conditional) way. However, since the update rule is a consequence of the error bound minimization process, the way these changes are really affecting the theoretical properties and optimality of AdaBoost cannot be guaranteed.

Considering those previous variations as \emph{heuristic}, Masnadi-Shirazi and Vasconcelos \cite{MasnadiVasconcelos11} proposed a theoretically sound approach based on the Statistical View of Boosting. According to this interpretation \cite{Friedman00} boosting algorithms can be seen as round-by-round estimations building an additive logistic regression model, and the exponential error bound can be modeled as the minimization of the next expression, where $E$ means expectation:

\begin{equation}
J(f)=E\left(e^{-yf(x_i)}\right)
\end{equation}

Setting the derivative ${\partial J(f)}/{\partial f(x)}$ to zero, we can obtain the solution of the minimization problem as the weighted logistic transform of $P(y=1|x)$ (\ref{stat_sol}).

\begin{equation}
\label{stat_sol}
\begin{split}
f(x)=\frac{1}{2}\log\frac{P\left(y=1|x\right)}{P\left(y=-1|x\right)}
\end{split}
\end{equation}

Following this perspective, Masnadi-Shirazi and Vasconcelos adapted it to the cost-sensitive case

\begin{equation}
J(f)=E\left(\llbracket y=1 \rrbracket e^{-C_{P}f(x_i)} + \llbracket y=-1 \rrbracket e^{C_{N}f(x_i)}\right)
\end{equation}

\begin{equation}
\label{stat_sol_asym}
\begin{split}
f(x)=\frac{1}{C_{P}+C_{N}}\log\frac{C_{P}P\left(y=1|x\right)}{C_{N}P\left(y=-1|x\right)}
\end{split}
\end{equation}

where $C_{P}$ and $C_{N}$ denote the misclassification costs for positives and negatives. The result of their derivation is the Cost-Sensitive AdaBoost algorithm we can see in algorithm \ref{csa_algorithm}. 

\begin{algorithm}[tb]
   \caption{Cost-Sensitive AdaBoost}
   \label{csa_algorithm}
   \scriptsize
   \begin{algorithmic}
   \STATE {\bfseries Input:} \\Training set of $n$ examples: $(\mathbf{x_i}, y_i)$, where $y_{i}=$ 
   $\left\{
   \begin{array}{ll}
   1 & \mbox{if $1 \leq i \leq m$},\\
   -1 & \mbox{if $m < i \leq n$}.
   \end{array} \right.$\\   
   Pool of $F$ weak classifiers: $h_{f}(\mathbf{x})$\\ Cost parameters: $C_{P}$, $C_{N}$\\ Number of rounds: $T$
   \vspace{4pt}
   \STATE {\bfseries Initialize:} \\ Uniform distribution of weights for each class: $D(i)=$
   $\left\{
   \begin{array}{ll}
   \frac{1}{2m} & \mbox{if $1 \leq i \leq m$},\\
   \frac{1}{2(n-m)} & \mbox{if $m < i \leq n$}.
   \end{array} \right.$\\   
   \FOR{$t=1$ {\bfseries to} $T$}
   \vspace{4pt}
   \STATE Calculate parameters:\\
   $\mathcal{T}_{P}=\sum_{i=1}^{m}D(i)$\\
   $\mathcal{T}_{N}=\sum_{i=m+1}^{n}D(i)$\\
   \vspace{4pt}
   \FOR{$f=1$ {\bfseries to} $F$}
   \vspace{4pt}
   \STATE Pick up $f^{th}$ weak classifier: $h_{f}(\mathbf{x})$.\\
   \vspace{4pt}
   \STATE Calculate parameters:\\
   $D(i)= \left\{
   \begin{array}{ll}
   \mathcal{B}=\sum_{i=1}^{m}D(i)\llbracket y_{i} \neq h_{f}(\mathbf{x_{i}})\rrbracket,\\
   \mathcal{D}=\sum_{i=m+1}^{n}D(i)\llbracket y_{i} \neq h_{f}(\mathbf{x_{i}})\rrbracket.
   \end{array} \right. $\\
   
   \vspace{4pt}
   \STATE Find $\alpha_{t,f}$ solving the next hyperbolic equation:\\
   $2C_{P}\mathcal{B}\cosh\left(C_{P}\alpha_{t,f}\right)+2C_{N}\mathcal{D}\cosh\left(C_{N}\alpha_{t,f}\right)=C_{1}\mathcal{T}_{P}e^{-C_{P}\alpha{t,f}}+C_{2}\mathcal{T}_{N}e^{-C_{N}\alpha_{t,f}}$\\
   \vspace{4pt}
   \STATE Compute the loss of the weak learner \\$L_{t,f}=\mathcal{B}\left(e^{C_{P}\alpha_{t,f}}-e^{-C_{P}\alpha_{t,f}}\right)+\mathcal{T}_{P}e^{-C_{P}\alpha_{t,f}}+\mathcal{D}\left(e^{C_{N}\alpha_{t,f}}-e^{-C_{N}\alpha_{t,f}}\right)+\mathcal{T}_{N}e^{-C_{N}\alpha_{t,f}}$
 	\vspace{4pt}
   \ENDFOR
   \vspace{4pt}
   \STATE Select the weak learner $\left(h_t(\mathbf{x}), \alpha_{t}(\mathbf{x})\right)$ of smallest loss in this round:  $\underset{f}{\operatorname{arg min}} \left[L_{t,f}\right]$\\
   \STATE Update weights: \\ 
   $D(i)=\left\{
   \begin{array}{ll}
   D(i)e^{-C_{P}\alpha_{t}h_{t}\left(\mathbf{x_i}\right)} & \mbox{if $1 \leq i \leq m$},\\
   D(i)e^{C_{N}\alpha_{t}h_{t}\left(\mathbf{x_i}\right)} & \mbox{if $m < i \leq n$}.
   \end{array} \right.$\\
   \vspace{4pt}
   \ENDFOR
   \vspace{4pt}
   \STATE {\bfseries Final Classifier:}\\
   \vspace{4pt}
    $H(\mathbf{x})=\mathrm{sign}\left(\sum_{t=1}^{T}\alpha_{t}h_{t}(\mathbf{x})\right)$
\end{algorithmic}
\end{algorithm}

It is important to note that, for the sake of homogeneity and simplicity, we have kept and followed the original notation by Schapire and Singer \cite{SchapireSinger99} along the entire paper. Because of this, we have had to adapt the notation used in \cite{MasnadiVasconcelos11} to this format. As well as in that work, we have also particularized our analysis to the most common case of having an initial pool of weak classifiers.

\section{AdaBoostDB}
\label{sec:AdaBoostDB}

Following the analytical guidelines proposed by Schapire and Singer \cite{SchapireSinger99}, in this section we will present and theoretically derive our asymmetric generalization of AdaBoost, AdaBoostDB, based on modifying the usual cost-insensitive exponential error bound with class-dependent bases.

\subsection{Double-Base Error Bound}
\label{sub_sec:Double}

Based on the inequality in equation (\ref{bound_ineq_eqn}), the original AdaBoost formulation is geared to minimize an exponential error bound $\tilde{E}_T$ over the weighted training error $E_{T}$ (\ref{exp_bound_eqn}). For minimization purposes, the specific exponential base $\beta$ is irrelevant whenever $\beta>1$ so, for simplicity, the selected base in the classical formulation of AdaBoost is $\beta=e$.

\begin{equation}
\label{exp_bound_eqn}
E_{T}=\sum_{i=1}^{n} D_{1}(i)\llbracket H(x_{i}) \neq y_{i}\rrbracket \leq \sum_{i=1}^{n} D_{1}(i)\beta^{ -y_{i} f(x_{i})}=\tilde{E}_{T}
\end{equation}

If we suppose, without loss of generality, that our training set is divided into two meaningful subsets (the first $m$ examples, positives, and the rest, negatives) we can define exponential bounds with different bases for each one. Calling $\beta_{P}$ and $\beta_{N}$ to these bases, the decomposed exponential bound $\tilde{E_{T}}$ can be expressed as equation (\ref{exp_bound_eqn_asym}). We will assume, without loss of generality, that $\beta_P \geq \beta_N >1$.

\begin{equation}
\label{exp_bound_eqn_asym}
E_{T} \leq \tilde{E}_T = \sum_{i=1}^{m} D_{1}(i){\beta_P}^{ -y_{i} f(x_{i})} + \sum_{i=m+1}^{n} D_{1}(i){\beta_N}^{ -y_{i} f(x_{i})}
\end{equation}

\begin{figure}[ht]
\begin{center}
\centerline{\includegraphics[width=3.5in]{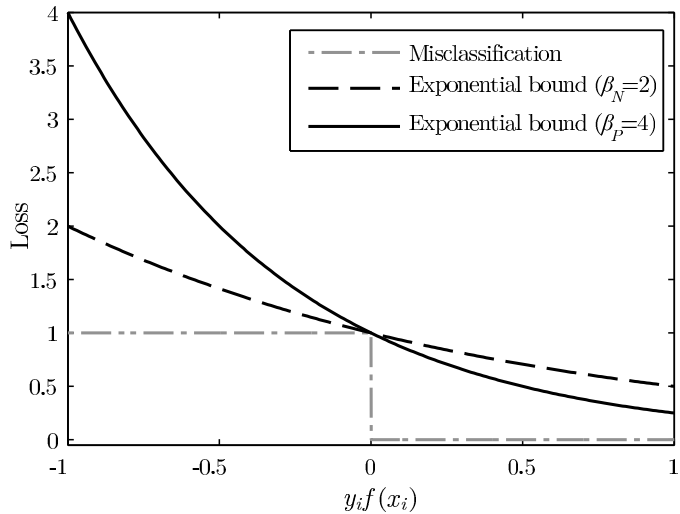}}
\caption{Misclassification and AdaBoost exponential training error bounds with different bases. The final score of the strong classifier is represented in the horizontal axis (negative sign for errors and positive for correct classifications), while vertical axis is the loss related to each possible score.}
\label{exp_bound_fig}
\end{center}
\end{figure}

This base-dependent behavior can be graphically analyzed in Figure~\ref{exp_bound_fig}: the greater one base is respect to the other, the more penalized are its respective errors. Therefore, associating $\beta_{P}$ to positive examples subclass and $\beta_{N}$ to the negative ones, this imbalanced behavior can be directly mapped to a class imbalanced cost-sensitive approach.

Rewriting the expression of ${\tilde{E}_{T}}$ (\ref{exp_bound_eqn_asym}) in terms of asymmetric exponents (\ref{exp_bound_csb}), this double-base perspective can be immediately linked with the Cost-Sensitive Boosting framework: both approaches are equivalently parameterized by class-conditional costs ($C_P=\log\left(\beta_P\right)$ and $C_N=\log\left(\beta_N\right)$, for positives and negatives respectively) and have the same statistical meaning (\ref{stat_sol}). 

\begin{equation}
\label{exp_bound_csb}
\begin{split}
\tilde{E}_T &= \sum_{i=1}^{m} D_{1}(i){e}^{ -\log{\beta_P} y_{i} f(x_{i})} + \sum_{i=m+1}^{n} D_{1}(i){e}^{ -\log{\beta_N} y_{i} f(x_{i})}\\ 
& = \sum_{i=1}^{m} D_{1}(i){e}^{ -C_P y_{i} f(x_{i})} + \sum_{i=m+1}^{n} D_{1}(i){e}^{ -C_N y_{i} f(x_{i})}
\end{split}
\end{equation}

\subsection{Algorithm Derivation}
\label{sub_sec:Derivation}

As we have just seen, the double-base approach shares with Cost-Sensitive AdaBoost a common theoretical root. However, our change in the point of view, along with a derivation inspired in the original framework by Schapire and Singer \cite{SchapireSinger99} (instead of the Statistical View of Boosting used to derive Cost-Sensitive Boosting), will allow us to follow a different derivation pathway, ending in a much more efficient formulation.

Let us suppose, again, that the first $m$ examples of the training set are positives and the rest are negatives, so the base-dependent behavior results in a class-dependent one. In this case, we can also split the initial weight distribution $D_{1}$ into two class-dependent \emph{subdistributions}, $D_{P,1}$ and $D_{N,1}$, for positives and negatives respectively:

\begin{gather}
\label{pos_weight_ini_eqn}
D_{P,1}(i)=
\frac{D_{1}(i)}{\sum_{i=1}^{m} D_{1}(i) }, \hspace{1em} \textrm{for } i=1,\ldots,m\\
\label{neg_weight_ini_eqn}
D_{N,1}(i)=
\frac{D_{1}(i)}{\sum_{i=m+1}^{n} D_{1}(i) }, \hspace{1em} \textrm{for } i=m+1,\ldots,n
\end{gather}

Defining the global weight of each class, $W_{P}$ and $W_{N}$ as follows, 

\begin{gather}
\label{pos_total_weight}
W_{P}=\sum_{i=1}^{m} D_{1}(i)\\
\label{neg_total_weight}
W_{N}=\sum_{i=m+1}^{n} D_{1}(i)
\end{gather}

the error bound $\tilde{E}_{T}$ can be decomposed into two class-dependent bounds $\tilde{E}_{P,T}$ and $\tilde{E}_{N,T}$. 

\begin{equation}
\label{asym_bound_eqn}
\begin{split}
\tilde{E}_{T}& =W_P\sum_{i=1}^{m} D_{P,1}(i){\beta_P}^{ -y_{i} f(x_{i})}+W_N\sum_{i=m+1}^{n} D_{N,1}(i){\beta_N}^{ -y_{i} f(x_{i})} = W_P\tilde{E}_{P,t}+W_N\tilde{E}_{N,t}
\end{split}
\end{equation}

Both error components are formally identical to the original bound (except for the weight distributions) allowing us to directly insert different exponential bases for each of them. This is just what we wanted. As in the original AdaBoost formulation, initial weight subdistributions can be extrapolated to round-by-round ones ($D_{P,t}$ and $D_{N,t}$) being iteratively updated and normalized in an analogous way\footnote{For shortness we will only show equations (\ref{posweight_rule_eqn}), (\ref{accpos_eqn}) and (\ref{exp_bound_pos_eqn}) for the positive class case. The negative ones are completely analogous to them.}.

\begin{equation}
\begin{split}
\label{posweight_rule_eqn}
D_{P,(t+1)}(i)&= \frac{D_{P,t}(i) {\beta_{P}}^{-\alpha_{t} y_{i} h_{t}(x_{i})}}{\sum_{i=1}^{m} D_{P,t}(i) {\beta_P}^{-\alpha_{t} y_{i} h_{t}(x_{i})}} = \frac{D_{P,t}(i) {\beta_{P}}^{-\alpha_{t} y_{i} h_{t}(x_{i})}}{Z_{P,t}}
\end{split}
\end{equation}

Two new parameters $A_{P/N,t}$ can also be defined as accumulators of the training behavior over each class until round $t$ (\ref{accpos_eqn}). Their definition can be obtained by unraveling the weight update rule, and allows us to decouple each class error bound into two factors (\ref{exp_bound_pos_eqn}): one only depending on the previous rounds $A_{P/N,t-1}$ and other depending on the performance of the current round $Z_{P/N,t}$ (with an homologous meaning to $Z_t$ in the original AdaBoost formulation).


\begin{equation}
\label{accpos_eqn}
A_{P,t}=
\prod_{k=1}^{t} Z_{P,k}=
\sum_{i=1}^{m} D_{P,1}(i) {\beta_P}^{-\sum_{k=1}^{t}\alpha_{k} y_{i} h_{k}(x_{i})}\\
\end{equation}

\begin{equation}
\label{exp_bound_pos_eqn}
\begin{split}
\tilde{E}_{P,t}&=\sum_{i=1}^{m} D_{P,1}(i){\beta_P}^{-\sum_{k=1}^{t}\alpha_{k} y_{i} h_{k}(x_{i})}=A_{P,t-1} \sum_{i=1}^{m} D_{P,t}(i){\beta_P}^{-y_{i}\alpha_{t}h_{t}(x_{i})}=A_{P,t-1}Z_{P,t}
\end{split}
\end{equation}

As a consequence, the total error to minimize ($\tilde{E_t}$) can be expressed as (\ref{exp_bound_asym_eqn}).

\begin{equation}
\label{exp_bound_asym_eqn}
\begin{split}
\tilde{E}_{t}&=W_{P}A_{P,t-1}Z_{P,t}+W_{N}A_{N,t-1}Z_{N,t}\\
&=W_{P}A_{P,t-1}\sum_{i=1}^{m}D_{P,t}(i){\beta_P}^{-y_{i}\alpha_{t}h_{t}(x_{i})}+W_{N}A_{N,t-1}\sum_{i=m+1}^{n}D_{N,t}(i){\beta_N}^{-y_{i}\alpha_{t}h_{t}(x_{i})}
\end{split}
\end{equation}

Due to the convexity of exponential functions, the minimum of this bound $\tilde{E}_{t}$ can be analytically found by canceling its derivative. Defining the cost parameters as commented in the previous section ($C_{P}=\log(\beta_{P})$ and $C_{N}=\log(\beta_{N})$), and bearing in mind equation (\ref{approx_exp_eqn})\footnote{Equation (\ref{approx_exp_eqn}) is strictly true for the discrete case, when weak hypothesis are 1 or -1. However, if weak hypothesis were real in the range $[-1,1]$, this equation would transform in an upper bound as explained in \cite{SchapireSinger99}. In that case we would be minimizing an upper bound on $\tilde{E}_{t}$ instead of $\tilde{E}_{t}$ directly, which is the same behavior as in the original AdaBoost with real-valued weak predictors.}, the goal derivative can be expressed as (\ref{deriv_eqn}).

\begin{equation}
\label{approx_exp_eqn}
\begin{split}
\beta^{-\alpha_{t}y_{i}h_{t}(x_i)} = \frac{1+y_{i}h_{t}(x_{i})}{2}\beta^{-\alpha_{t}} + \frac{1-y_{i}h_{t}(x_{i})}{2}\beta^{\alpha_{t}}\\
\end{split}
\end{equation}

\begin{equation}
\label{deriv_eqn}
\begin{split}
\frac{\partial\tilde{E}_{t}}{\partial\alpha_{t}}&= C_{P}W_{P}A_{P,t-1}\sum_{\textrm{Pos nok}}D_{P,t}(i){\beta_P}^{\alpha_{t}}-C_{P}W_{P}A_{P,t-1}\sum_{\textrm{Pos ok}}D_{P,t}(i){\beta_P}^{-\alpha_{t}}\\
&\quad+C_{N}W_{N}A_{N,t-1}\sum_{\textrm{Neg nok}}D_{N,t}(i){\beta_N}^{\alpha_{t}}-C_{N}W_{N}A_{N,t-1}\sum_{\textrm{Neg ok}}D_{N,t}(i){\beta_N}^{-\alpha_{t}}=0
\end{split}
\end{equation}

Since $C_{P}$ and $C_{N}$ do not have to be integer values in general, the real asymmetry only relies on their relative magnitudes (how much a positive costs over a negative), so we will always find equivalent integer values to play this role whatever the desired asymmetry is.

At this point, with $\alpha_t$ as unknown variable, the minimization equation can be modeled as a polynomial (\ref{polin_eqn}) by making a change of variable (\ref{coeff_eqn1}) and rewriting it in terms of parameters (\ref{coeff_eqn2}, \ref{coeff_eqn3}, \ref{coeff_eqn4}, \ref{coeff_eqn5}), instead of the hyperbolic model used in \cite{MasnadiVasconcelos11}.

\begin{gather}
\label{coeff_eqn1}
x=e^{\alpha_{t}}\\
\label{coeff_eqn2}
a=\frac{C_{P}W_{P}A_{P,t-1}}{C_{P}W_{P}A_{P,t-1}+C_{N}W_{N}A_{N,t-1}}\\
\label{coeff_eqn3}
b=\frac{C_{N}W_{N}A_{N,t-1}}{C_{P}W_{P}A_{P,t-1}+C_{N}W_{N}A_{N,t-1}}\\
\label{coeff_eqn4}
\varepsilon_{P,t}=\sum_{\textrm{Pos nok}}D_{P,t}(i)\\
\label{coeff_eqn5}
\varepsilon_{N,t}=\sum_{\textrm{Neg nok}}D_{N,t}(i)
\end{gather}

\begin{equation}
\label{polin_eqn}
a\cdot\varepsilon_{P,t}\cdot x^{2C_{P}}+b\cdot\varepsilon_{N,t}\cdot x^{C_{P}+C_{N}}-b\left(1-\varepsilon_{N,t}\right)x^{C_{P}-C_{N}}-a\left(1-\varepsilon_{P,t}\right)=0
\end{equation}


The latter equation, where $x$ is the independent variable, has in general $2C_{P}$ possible solutions, from which, by the nature of the problem, we are only interested in those real and positive. It is easy to see that $a$, $b$, $\varepsilon_{P,t}$ and $\varepsilon_{N,t}$ are, by definition, all real values in the $[0,1]$ interval. As a consequence, there is only one sign change between consecutive coefficients of the polynomial, and by the Descartes' Rule of Signs we can ensure that the equation has \emph{only one real and positive solution} which is \emph{our solution}. 

The straightforward way to solve the posed problem is calculating the zeros of the polynomial to finally keep the only real and positive root. This process should be repeated for all the possible weak hypothesis in order to finally select that leading to the greatest \emph{goodness} $\alpha_{t}=\log(x_{root})$, that is, the one with the greatest root. This direct mechanism, requiring a scalar search, is very similar to that proposed in Cost-Sensitive AdaBoost but with the computational advantage of evaluating a polynomial instead of a hyperbolic function.

\subsection{Conditional Search}
\label{sub_sec:Conditional}

The main drawback of the straightforward solution in Section \ref{sub_sec:Derivation} is that it still requires the search of the associated root for every classifier in every boosting round. This could be very expensive in computational burden terms, for example, in applications needing to select from hundreds of thousands different classifiers evaluated over several thousands of training examples such as computer vision algorithms \cite{ViolaJones04}. Nevertheless, a slight change in the point of view can serve to drastically reduce this computational burden. If we define functions $V(x)$ and $S(x)$ as follows, we can rewrite equation (\ref{polin_eqn}) as $S(x)=V(x)$.

\begin{gather}
\label{graph_eqn}
S(x)= a+b\cdot x^{C_{P}-C_{N}}\\ 
V(x)= a\cdot\varepsilon_{P,t}\left(x^{2C_{P}}+1\right)+b\cdot\varepsilon_{N,t}\left(x^{C_{P}+C_{N}}+x^{C_{P}-C_{N}}\right)
\end{gather}

The first function $S(x)$ is a polynomial whose coefficients are parameters $a$ and $b$, which only depend on the previous boosting rounds. The second one, $V(x)$, has coefficients also depending on $\varepsilon_{P,t}$ and $\varepsilon_{N,t}$ so it has a dependence with the current round as well. As a result, the minimization procedure of a given round can be modeled as the crossing point between a \emph{static} function $S(x)$, fixed for the current round, and a \emph{variable} function $V(x)$. 

\begin{figure}[ht]
\begin{center}
\centerline{\includegraphics[width=\columnwidth]{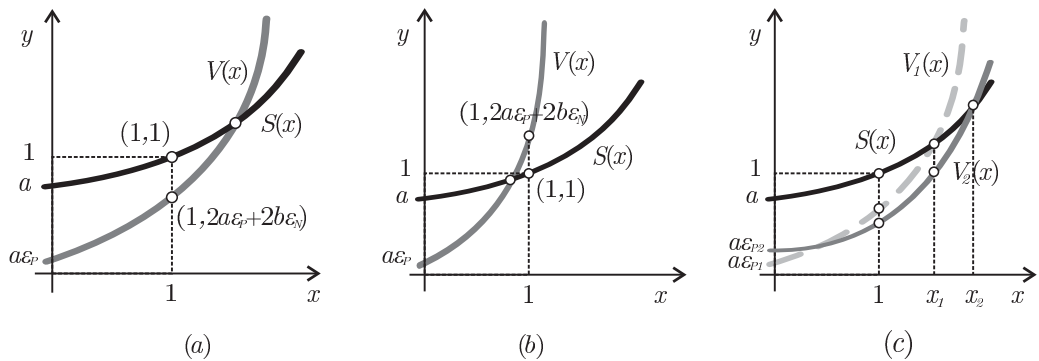}}
\caption{(a) Crossing point scenario for the \emph{static} $S(x)$ and \emph{variable} $V(x)$ functions (a) modeling the minimization problem. Graphical representation of the \emph{Contribution} (not fulfilled) (b) and \emph{Improvement} (c) conditions.}
\label{crossing_scenario}
\end{center}
\end{figure}

It is important also to bear in mind some specificities (the problem is graphically shown in Figure~\ref{crossing_scenario}):

\begin{itemize}
\item By definition all parameters $a$, $b$, $\varepsilon_{P,t}$ and $\varepsilon_{N,t}$ are positives, so both functions are increasing for $x\geq0$.
\item The crossing points with the y-axis are $(0,a\cdot\varepsilon_{P,t})$ for $V(x)$, and $(0,a)$ for $S(x)$. Taking into account that $\varepsilon_{P,t}\leq1$ we have $V(0)\leq S(0)$.
\item When $x\to\infty$, $V(x)>S(x)$.
\item There is only one positive crossing point.
\end{itemize}

Descartes' rule of signs ensures us the existence of one crossing point, but only solutions satisfying $x>1$ are interesting for the classification problem: only weak hypothesis with \emph{some goodness}, i.e.\ $\alpha_{t}>0$, are really contributing for the strong classifier. This \emph{Contribution Condition} can be formalized as follows (\ref{contrib_cond}), and any weak classifier that does not meet this requirement should be directly discarded for the current round without more computation.

\begin{equation}
\label{contrib_cond}
V(1)<S(1)=1 \Rightarrow a\cdot\varepsilon_{P}+b\cdot\varepsilon_{N} < \frac{1}{2}
\end{equation}

On the other hand, once we have computed a valid solution, to comparatively evaluate any other candidate we just need to know if its related root (i.e. its goodness $\alpha_t$) is greater to the one we already have. Using this information, we would only have to effectively calculate the specific root (i.e. run the scalar search) for those weak classifiers with greater roots, directly rejecting the other ones. Bearing in mind that both $V(x)$ and $S(x)$ are increasing functions, given two possible weak classifiers with their associated functions $V_{1}(x)$ and $V_{2}(x)$ and the solution $x_1$ for the first of them, the second classifier will only be better than the first one if $V_1(x_1)>V_2(x_1)$. We will call this rule as the \emph{Improvement Condition}.

Applying both conditions to the weak hypothesis searching process in a nested way, the average number of zeros effectively computed decreases over 99.5\% with respect to the straightforward solution, while consuming only 0.41\% of its time (more details in section \ref{sec:time_cmp}). It is important to emphasize that this improved searching technique, which we have coined as \emph{Conditional Search}, and the huge computational saving it brings, is made possible by the polynomial and double-base modeling of the proposed framework.

A compact summary (for a direct implementation) of the final version of AdaBoostDB algorithm, including the Conditional Search, is given in Algorithm~\ref{abc_algorithm}.

\begin{algorithm}[tb]
   \caption{AdaBoostDB}
   \label{abc_algorithm}
   \scriptsize
\begin{algorithmic}
   \STATE {\bfseries Input:} \\Training set of $n$ examples: $(\mathbf{x_i}, y_i)$, where $y_{i}= \left\{
   \begin{array}{cl}
   1 & \mbox{if $1 \leq i \leq m$},\\
   -1 & \mbox{if $m < i \leq n$}.
   \end{array} \right. $\\
Distribution of associated weights: $D(i)$\\ Pool of $F$ weak classifiers: $h_{f}(\mathbf{x})$\\ Cost parameters: $C_{P}$, $C_{N}$\\ Number of rounds: $T$\\
\vspace{4pt}
   \STATE {\bfseries Initialize:} \\Weight subdistributions:
   $\left\{
   \begin{array}{ll}
   D_{P}(i)=\frac{D(i)}{\sum_{i=1}^{m}D(i)} & \mbox{if $1 \leq i \leq m$},\\  			   
   D_{N}(i)=\frac{D(i)}{\sum_{i=m+1}^{n}D(i)} & \mbox{if $m < i \leq n$}.
   \end{array} \right.$\\
   Accumulators: $A_{P}=1$, $A_{N}=1$.\\
   \vspace{4pt}
   \FOR{$t=1$ {\bfseries to} $T$}
   \vspace{4pt}
   \STATE {\bfseries Initialize:} \\Minimum root: $r=1$\\ Minimum root vector: $\vec{r}=(2,2)$\\ Scalar product: $s=1$
      \vspace{4pt}
   \STATE Update accumulators:
   $\left\{
   \begin{array}{ll}
   A_{P}=A_{P}\sum_{i}D_{P}(i),\\
   A_{N}=A_{N}\sum_{i}D_{N}(i).
   \end{array} \right.$\\

   \STATE Normalize weight subdistributions:
   $\left\{
   \begin{array}{ll}
    D_{P}(i)=\frac{D_{P}(i)}{\sum_{i=1}^{m}D_{P}(i)},\\
   D_{N}(i)=\frac{D_{N}(i)}{\sum_{i=m+1}^{n}D_{N}(i)}.
   \end{array} \right.$\\
   
   \STATE Calculate static parameters: 
   $\left\{
   \begin{array}{ll}
   a=\frac{C_{P}A_{P}}{C_{P}A_{P}+C_{N}A_{N}},\\
   b=\frac{C_{N}A_{N}}{C_{P}A_{P}+C_{N}A_{N}}.
   \end{array} \right.$\\
   
      \vspace{4pt}
   \FOR{$f=1$ {\bfseries to} $F$}
      \vspace{4pt}
   \STATE Calculate variable parameters:
   $\left\{
   \begin{array}{ll}
   \varepsilon_{P,f}=\sum_{i=1}^{m}D_{P}(i)\llbracket y_{i} \neq h_{f}(\mathbf{x_{i}})\rrbracket,\\
   \varepsilon_{N,f}=\sum_{i=m+1}^{n}D_{N}(i)\llbracket y_{i} \neq h_{f}(\mathbf{x_{i}})\rrbracket.
   \end{array} \right.$\\
   \vspace{4pt}
   Calculate current classifier vector: $\vec{c}=(a\cdot\varepsilon_{P,f},b\cdot\varepsilon_{N,f})$\\
   \vspace{4pt}
   
 \fbox
 {\parbox{12cm} {
   \textbf{CONDITIONAL SEARCH}
   \vspace{2pt}
	 \IF {$a\cdot\varepsilon_{P,f}+b\cdot\varepsilon_{N,f}<\frac{1}{2}$ [\textbf{Contribution Condition}]}
   \vspace{4pt}
   \IF {$\vec{c}\cdot\vec{r}>s$ [\textbf{Improvement Condition}]}
   \vspace{4pt}
   \STATE Search the only real and positive root $r$ of  the polynomial:\\
    $(a\cdot\varepsilon_{P,f})x^{2C_{P}}+ (b\cdot\varepsilon_{N,f})x^{C_{P}+C_{N}}+b(\varepsilon_{N,f}-1)x^{C_{P}-C_{N}}+a(\varepsilon_{P,f}-1)=0$
    \vspace{4pt}
    
   \STATE Update parameters: 
   $\left\{
   \begin{array}{ll}
   \vec{r}=(r^{2C_{P}+1},r^{C_{P}+C_{N}}+r^{C_{P}+C_{N}}),\\
   s=\vec{c}\cdot\vec{r}.
   \end{array} \right.$\\
   
   \vspace{2pt}
   \STATE Keep $h_{f}(i)$ as round $t$ solution.
   \vspace{4pt}
   \ENDIF
   \vspace{4pt}
   \ENDIF}
}
   
   \vspace{4pt}
   \ENDFOR
   \vspace{4pt}
   \STATE Calculate goodness parameter: $\alpha_{t}=\log{(r)}$
   \vspace{4pt}
   \STATE Update weights subdistributions: 
   $\left\{
   \begin{array}{ll}
   D_{P}(i)=D_{P}(i)\exp(-C_{P}\alpha_{t}h_{t}(i)),\\
   D_{N}(i)=D_{N}(i)\exp(C_{N}\alpha_{t}h_{t}(i)).
   \end{array} \right.$\\
   \vspace{4pt}
   \ENDFOR
   \vspace{4pt}
   \STATE {\bfseries Final Classifier:}\\ 
   \vspace{4pt}
   $H(x)=\mathrm{sign}\left( \sum_{t=1}^{T}\alpha_{t}h_{t}(x)\right)$
\end{algorithmic}
\end{algorithm}


\section{Experiments}
\label{sec:Experiments}

To show and assess the performance of AdaBoostDB in practical terms we have conducted a series of empirical experiments to analyze the asymmetric behavior of the algorithm, comparing it with theoretical optimal classifiers and with Cost-Sensitive AdaBoost using synthetic and real datasets.

\subsection{Experimental Framework}
\label{sub_sec:Exper_Framework}

Cost-sensitive classification problems can be totally parameterized in terms of a cost matrix \cite{Elkan01}, whose components are the costs related to each possible decision. For a two-class problem this matrix can be expressed as follows:

\begin{equation}
\begin{tabular} {l c c c c}
& & Negative & Positive &\\
Classified as Negative  &\multirow{2}{*}{{\Huge(}} & $c_{nn}$ & $c_{np}$ & \multirow{2}{*}{{\Huge)}}\\
Classified as Positive & & $c_{pn}$ & $c_{pp}$ &
\end{tabular}
\end{equation}

In detection problems costs related to good decisions are considered null ($c_{nn}=c_{pp}=0$), so the cost matrix is only dependent on the two error-related parameters, $c_{np}$ and $c_{pn}$, which are directly assimilable to $C_P$ and $C_N$ in our previous theoretical analysis. Bearing in mind that the optimal decision is unchanged when the cost matrix is multiplied by a constant, the resulting matrix actually has only one degree of freedom, which we will call the \emph{asymmetry} ($\gamma$) of the problem.

\begin{equation}
\label{gamma}
\gamma=\frac{c_{np}}{c_{np}+c_{pn}}=\frac{C_{P}}{C_{P}+C_{N}}
\end{equation}

The traditional way to evaluate and compare the behavior of different classifiers across different working points has been based on the analysis of ROC curves \cite{Provost97,Fawcett06}. Nevertheless, an alternative representation proposed by C. Drummond and R.C. Holte \cite{Drummond00}, dual respect to traditional ROC curves and based on expected costs, has been shown to be more appropriate for cost-sensitive classification problems (cost is explicitly presented, enabling direct visual interpretations and comparisons). Our experimental analysis is based on these representations.

Following guidelines in \cite{Drummond00}, the \emph{Probability Cost Function} (PCF) and the \emph{Normalized Expected Cost} (NEC) are defined in equations \ref{PCF} and \ref{NEC}, where $p(+)$ and $p(-)$ are the prior probabilities of an example to be positive or negative, while $FN$ and $FP$ are, respectively, the false negatives and false positives rates obtained by the classifier.

\begin{gather}
\label{PCF}
PCF=\frac{p(+)C_{P}}{p(+)C_{P}+p(-)C_{N}}\\
\label{NEC}
NEC=FN \cdot PCF+FP
\end{gather}

\subsection{Bayes Error Rates}
\label{sub_sec:bayes_risk}

\begin{figure}[ht]
\begin{center}
\centerline{\includegraphics[width=\columnwidth]{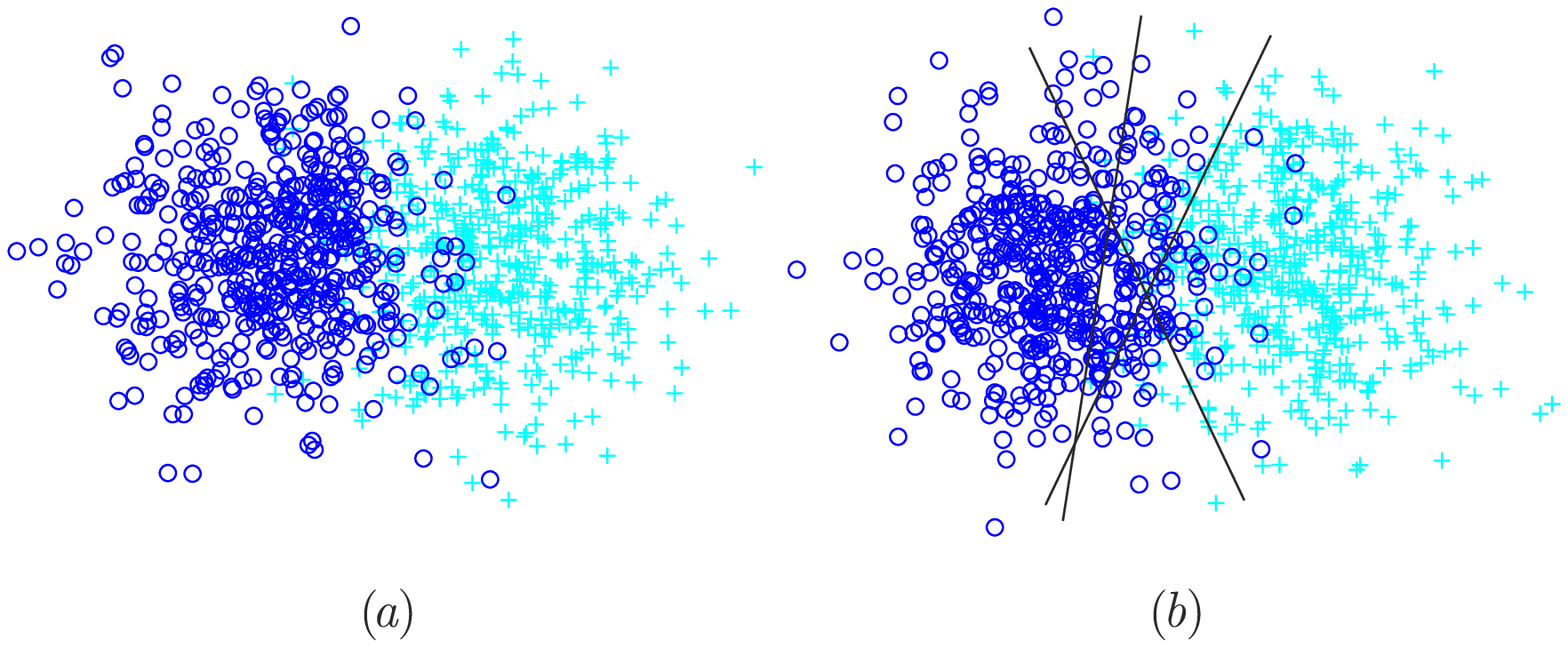}}
\caption{\emph{Bayes Risk} datasets used in our experiments: Positive examples are marked as `$+$', while negatives are `$\circ$'. In figure b examples of weak classifiers are shown.}
\label{bayes_dataset}
\end{center}
\end{figure}

As first step, we are going to compare AdaBoostDB classifiers with their optimal Bayes classifiers counterparts for different cost combinations. To this end, we have defined a synthetic dataset scenario from which we can easily calculate the theoretically optimal classifier following the Bayes Risk Rule. This synthetic scenario is illustrated in Figure \ref{bayes_dataset}: Two bivariate normal point clouds, one for positives and one for negatives, with the same priors and variances but different means. As customary in many boosting works \cite{SchapireSinger99,ViolaJones04,MasnadiVasconcelos11} weak learners are stumps (the quintessential weak classifier) computed in this case, over the projection of the points on a discrete range of angles in the 2D space (Figure \ref{bayes_dataset}b).

Two different random datasets were generated, one for training and the other one for test. Nineteen different asymmetries to evaluate have also been defined, trying to sweep a wide range of cost combinations: 

\begin{equation}
\label{cost_comb}
\begin{split}
(C_P, C_N)\in \{ & (1,100), (1,50), (1,25), (1,10), (1,7), (1,5), (1,3), (1,2), (2,3), (1,1),\\ 
& (3,2), (2,1), (3,1), (5,1), (7,1), (10,1), (25,1), (50,1), (100,1)\}
\end{split}
\end{equation}

Therefore, 19 different AdaBoostDB classifiers were trained to be compared with their respective optimal Bayes classifiers counterparts, over the same test set. In addition, another 19 classifiers using Cost-Sensitive AdaBoost have also been trained as a preliminary comparative between this algorithm and AdaBoostDB (we will delve in this issue in section \ref{subsec:ADB_vs_CSB}).

\begin{figure}[ht]
\begin{center}
\centerline{\includegraphics[width=5in]{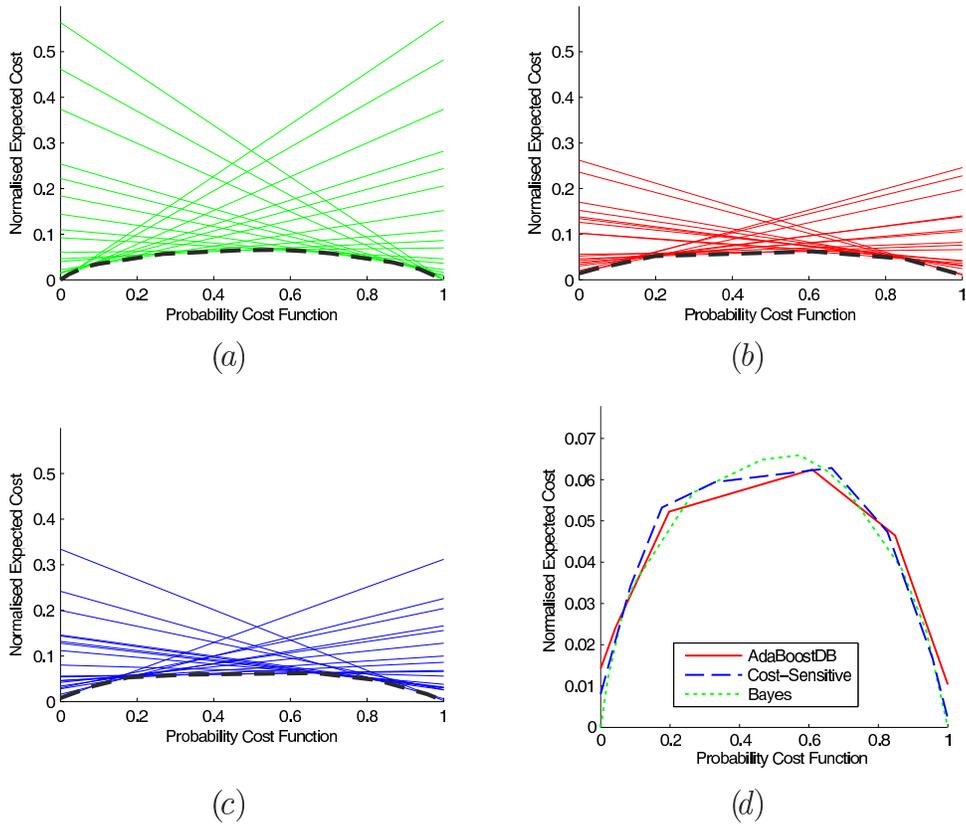}}
\caption{Lower envelope graphic representations for the three classifiers families: (a) Bayes, (b) AdaBoostDB and (c) Cost-Sensitive AdaBoost. Figure (c) shows the three lower envelopes superimposed.}
\label{envelopes_graphic}
\end{center}
\end{figure}

The goal of our first comparative test (also based on \cite{Drummond00}) is to compute the lower envelope of each set of classifiers (Bayes, AdaBoostDB and Cost-Sensitive AdaBoost) in the cost space. This cost space is defined by the relationship between Probability Cost Function (x-axis) and Normalized Expected Cost (y-axis). In this framework, every classifier, though trained for a specific asymmetry, can be tested in arbitrary cost scenarios (different asymmetries for the same test set) thus drawing a line passing by  $(0, FP)$ and $(1, FN)$ in the cost space. As a result, each family of classifiers will be represented by a collection of lines whose lower envelope defines the minimum cost classifier along the operating range (see Figure \ref{envelopes_graphic}). Comparing the three resulting lower envelopes (Figure \ref{envelopes_graphic}c) we can appreciate an equivalent behavior with only slight differences among them.

\begin{figure}[ht!]
\begin{center}
\centerline{\includegraphics[width=5in]{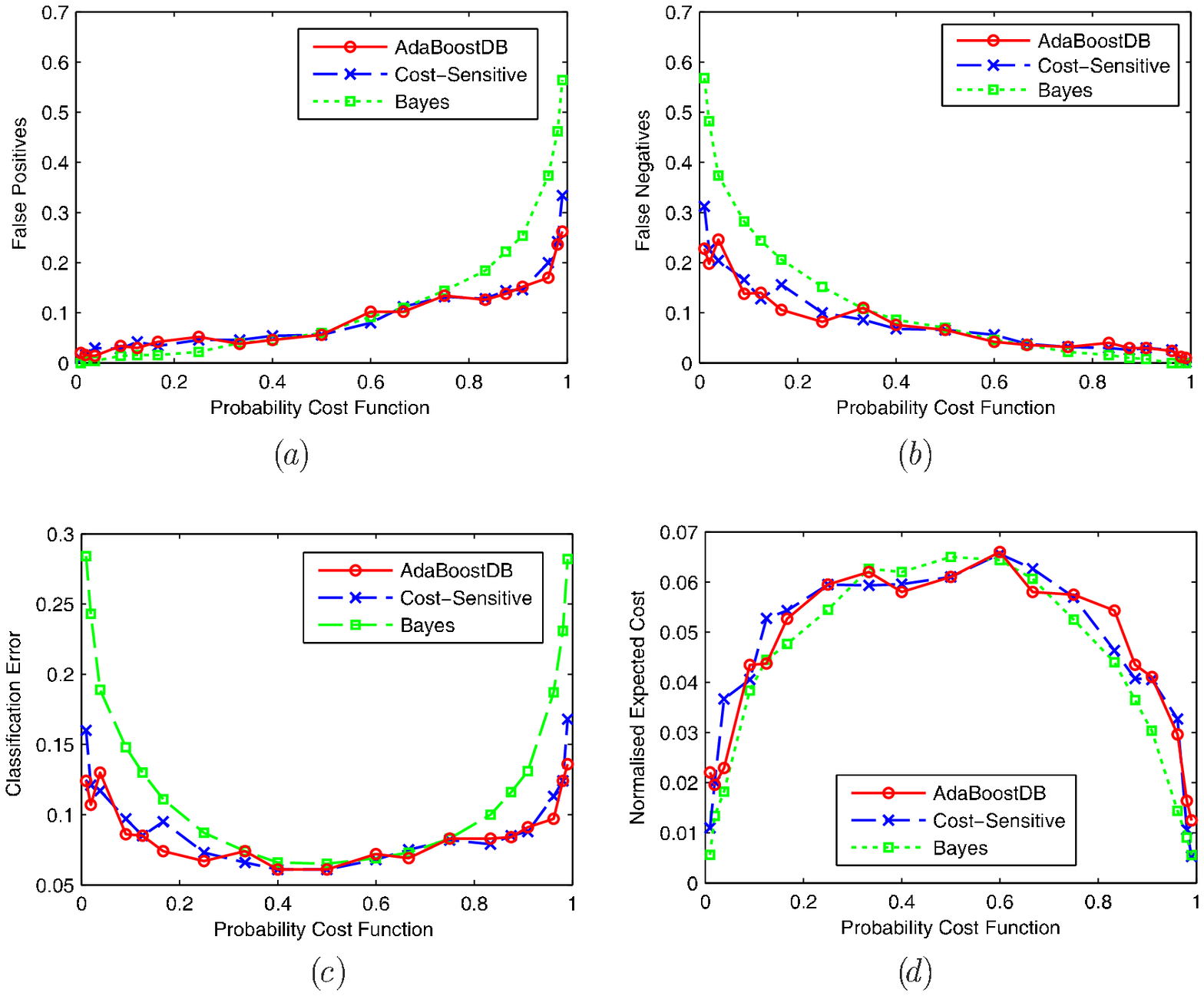}}
\caption{Performance comparison of classifiers obtained by AdaBoostDB, Bayes and Cost-Sensitive AdaBoost for each specific asymmetry over the Bayes synthetic test set. (a) False Positives, (b) False Negatives, (c) Classification Error, (d) Normalized Expected Cost.}
\label{bayes_perform_comparison}
\end{center}
\end{figure}

The second comparative test is among the same classifiers when tested for the specific asymmetry they were trained for. Results can be seen in Figure \ref{bayes_perform_comparison}. AdaboostDB performance follows the trend set by the Bayes optimal classifier, describing a consistent and gradual asymmetric behavior across the different costs and all the studied parameters (false positives, false negatives, classification error and normalized expected cost). Moreover, as we will comment in section \ref{subsec:ADB_vs_CSB}, the behavior of AdaBoostDB and Cost-Sensitive AdaBoost is virtually the same.

\subsection{Asymmetric behavior}
\label{sub_sec:Asymm_behavior}

Now the goal is to test the asymmetric behavior of AdaBoostDB over heterogeneous classification problems, using synthetic and real datasets and different cost requirements. 

\begin{figure}[ht]
\begin{center}
\centerline{\includegraphics[width=3.5in]{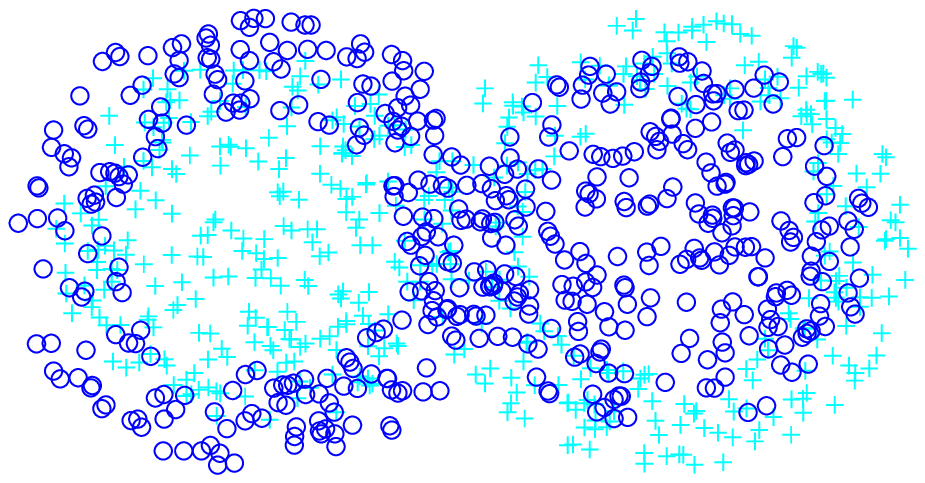}}
\caption{\emph{Two Clouds} dataset used in our experiments. Positive examples are marked as `$+$', while `$\circ$' are the negative ones (note that positive and negative classes are overlapped in both cases).}
\label{twoclouds_dataset}
\end{center}
\end{figure}

\emph{Synthetic datasets}: In addition to the dataset used in the last section (called as ``Bayes'' dataset), we will also use a two cloud scenario inspired by \cite{ViolaJones02}, in which positives and negatives are uniformly distributed in overlapping circular/annular regions with different centroids (see Figure \ref{twoclouds_dataset}). Features are again the projections of the examples over a discrete range of angles in the 2D space.

\emph{Real datasets}: We selected several datasets, asymmetric on their own definition, from UCI Machine Learning Repository \cite{UCIRepository} (Credit, Ionosphere, Diabetes and Spam). We have considered as positives the more valuable classes according to the original problems.

In both synthetic and real cases, weak learners are stumps. For every dataset and every cost requirement, we have followed a \emph{3-fold cross-validation} strategy to evaluate the asymmetric performance: the whole dataset is divided in three subsets, leaving iteratively one of them as test set and the other two forming the training set. As a result, for every dataset-cost combination, we can obtain the performance averages of the three classifiers.

Obtained results are shown in Table~\ref{asym_perform_tab}: As expected, when positives become more costly than negatives, false negative rates (FN, error in positives) tend to decrease while false positives rates (FP, error in negatives) tend to increase. In the opposite situation (when negatives become more costly than positives) roles are accordingly exchanged, showing a progressive and consistent asymmetric behavior, generalized across all the datasets and cost combinations. Information in the table is supplemented with two global performance measures, Classification Error (CE) and Normalized Expected Cost (NEC), of each experiment.

\begin{table*}[htbp]
  \centering
  \footnotesize
  \caption{{AdaBoostDB asymmetric behavior (false negatives, false positives, classification error and normalized expected cost) for each cost combination over the synthetic and UCI datasets.}}
  \scalebox{0.6}{
    \begin{tabular}{|c|c|c|c|c|c|c|c|c|}
    \hline
    \multirow{2}[4]{*}{\textbf{Cost}} & \multicolumn{4}{c|}{\textbf{Bayes}} & \multicolumn{4}{c|}{\textbf{TwoClouds}} \\
\cline{2-9}          & \textbf{FN} & \textbf{FP} & \textbf{CE} & \textbf{NEC} & \textbf{FN} & \textbf{FP} & \textbf{CE} & \textbf{NEC} \\
    \hline
    \textbf{[1,100]} & 2.13$\cdot10^{-1}$ & 3.21$\cdot10^{-2}$ & 1.22$\cdot10^{-1}$ & 3.39$\cdot10^{-2}$ & 9.12$\cdot10^{-1}$ & 6.02$\cdot10^{-3}$ & 4.59$\cdot10^{-1}$ & 1.50$\cdot10^{-2}$ \\
    \textbf{[1,50]} & 1.73$\cdot10^{-1}$ & 4.02$\cdot10^{-2}$ & 1.06$\cdot10^{-1}$ & 4.28$\cdot10^{-2}$ & 9.12$\cdot10^{-1}$ & 6.02$\cdot10^{-3}$ & 4.59$\cdot10^{-1}$ & 2.38$\cdot10^{-2}$ \\
    \textbf{[1,25]} & 1.73$\cdot10^{-1}$ & 3.21$\cdot10^{-2}$ & 1.02$\cdot10^{-1}$ & 3.75$\cdot10^{-2}$ & 9.12$\cdot10^{-1}$ & 6.02$\cdot10^{-3}$ & 4.59$\cdot10^{-1}$ & 4.09$\cdot10^{-2}$ \\
    \textbf{[1,10]} & 1.37$\cdot10^{-1}$ & 4.42$\cdot10^{-2}$ & 9.04$\cdot10^{-2}$ & 5.26$\cdot10^{-2}$ & 8.49$\cdot10^{-1}$ & 6.02$\cdot10^{-3}$ & 4.28$\cdot10^{-1}$ & 8.27$\cdot10^{-2}$ \\
    \textbf{[1,7]} & 1.37$\cdot10^{-1}$ & 4.02$\cdot10^{-2}$ & 8.84$\cdot10^{-2}$ & 5.22$\cdot10^{-2}$ & 7.85$\cdot10^{-1}$ & 2.21$\cdot10^{-2}$ & 4.04$\cdot10^{-1}$ & 1.17$\cdot10^{-1}$ \\
    \textbf{[1,5]} & 1.29$\cdot10^{-1}$ & 4.02$\cdot10^{-2}$ & 8.43$\cdot10^{-2}$ & 5.49$\cdot10^{-2}$ & 7.35$\cdot10^{-1}$ & 2.21$\cdot10^{-2}$ & 3.79$\cdot10^{-1}$ & 1.41$\cdot10^{-1}$ \\
    \textbf{[1,3]} & 1.20$\cdot10^{-1}$ & 3.61$\cdot10^{-2}$ & 7.83$\cdot10^{-2}$ & 5.72$\cdot10^{-2}$ & 7.43$\cdot10^{-1}$ & 3.82$\cdot10^{-2}$ & 3.91$\cdot10^{-1}$ & 2.14$\cdot10^{-1}$ \\
    \textbf{[1,2]} & 1.29$\cdot10^{-1}$ & 3.61$\cdot10^{-2}$ & 8.23$\cdot10^{-2}$ & 6.69$\cdot10^{-2}$ & 5.96$\cdot10^{-1}$ & 9.04$\cdot10^{-2}$ & 3.43$\cdot10^{-1}$ & 2.59$\cdot10^{-1}$ \\
    \textbf{[2,3]} & 1.04$\cdot10^{-1}$ & 4.82$\cdot10^{-2}$ & 7.63$\cdot10^{-2}$ & 7.07$\cdot10^{-2}$ & 4.78$\cdot10^{-1}$ & 1.81$\cdot10^{-1}$ & 3.29$\cdot10^{-1}$ & 3.00$\cdot10^{-1}$ \\
    \textbf{[1,1]} & 5.62$\cdot10^{-2}$ & 7.63$\cdot10^{-2}$ & 6.63$\cdot10^{-2}$ & 6.63$\cdot10^{-2}$ & 3.92$\cdot10^{-1}$ & 2.93$\cdot10^{-1}$ & 3.42$\cdot10^{-1}$ & 3.42$\cdot10^{-1}$ \\
    \textbf{[3,2]} & 6.02$\cdot10^{-2}$ & 7.63$\cdot10^{-2}$ & 6.83$\cdot10^{-2}$ & 6.67$\cdot10^{-2}$ & 2.23$\cdot10^{-1}$ & 4.08$\cdot10^{-1}$ & 3.15$\cdot10^{-1}$ & 2.97$\cdot10^{-1}$ \\
    \textbf{[2,1]} & 3.61$\cdot10^{-2}$ & 8.43$\cdot10^{-2}$ & 6.02$\cdot10^{-2}$ & 5.22$\cdot10^{-2}$ & 1.24$\cdot10^{-1}$ & 5.48$\cdot10^{-1}$ & 3.36$\cdot10^{-1}$ & 2.66$\cdot10^{-1}$ \\
    \textbf{[3,1]} & 4.42$\cdot10^{-2}$ & 8.84$\cdot10^{-2}$ & 6.63$\cdot10^{-2}$ & 5.52$\cdot10^{-2}$ & 4.82$\cdot10^{-2}$ & 6.53$\cdot10^{-1}$ & 3.50$\cdot10^{-1}$ & 1.99$\cdot10^{-1}$ \\
    \textbf{[5,1]} & 5.62$\cdot10^{-2}$ & 7.63$\cdot10^{-2}$ & 6.63$\cdot10^{-2}$ & 5.96$\cdot10^{-2}$ & 1.00$\cdot10^{-2}$ & 8.13$\cdot10^{-1}$ & 4.12$\cdot10^{-1}$ & 1.44$\cdot10^{-1}$ \\
    \textbf{[7,1]} & 4.82$\cdot10^{-2}$ & 1.04$\cdot10^{-1}$ & 7.63$\cdot10^{-2}$ & 5.52$\cdot10^{-2}$ & 1.20$\cdot10^{-2}$ & 8.61$\cdot10^{-1}$ & 4.37$\cdot10^{-1}$ & 1.18$\cdot10^{-1}$ \\
    \textbf{[10,1]} & 4.42$\cdot10^{-2}$ & 1.24$\cdot10^{-1}$ & 8.43$\cdot10^{-2}$ & 5.15$\cdot10^{-2}$ & 1.00$\cdot10^{-2}$ & 8.73$\cdot10^{-1}$ & 4.42$\cdot10^{-1}$ & 8.85$\cdot10^{-2}$ \\
    \textbf{[25,1]} & 3.21$\cdot10^{-2}$ & 2.05$\cdot10^{-1}$ & 1.18$\cdot10^{-1}$ & 3.88$\cdot10^{-2}$ & 1.00$\cdot10^{-2}$ & 9.48$\cdot10^{-1}$ & 4.79$\cdot10^{-1}$ & 4.61$\cdot10^{-2}$ \\
    \textbf{[50,1]} & 2.81$\cdot10^{-2}$ & 1.81$\cdot10^{-1}$ & 1.04$\cdot10^{-1}$ & 3.11$\cdot10^{-2}$ & 1.00$\cdot10^{-2}$ & 9.48$\cdot10^{-1}$ & 4.79$\cdot10^{-1}$ & 2.84$\cdot10^{-2}$ \\
    \textbf{[100,1]} & 2.81$\cdot10^{-2}$ & 1.85$\cdot10^{-1}$ & 1.06$\cdot10^{-1}$ & 2.97$\cdot10^{-2}$ & 1.00$\cdot10^{-2}$ & 9.48$\cdot10^{-1}$ & 4.79$\cdot10^{-1}$ & 1.93$\cdot10^{-2}$ \\
    \hline
    \multicolumn{1}{c}{} & \multicolumn{1}{c}{} & \multicolumn{1}{c}{} & \multicolumn{1}{c}{} & \multicolumn{1}{c}{} & \multicolumn{1}{c}{} & \multicolumn{1}{c}{} & \multicolumn{1}{c}{} & \multicolumn{1}{c}{} \\
    \hline
    \multirow{2}[4]{*}{\textbf{Cost}} & \multicolumn{4}{c|}{\textbf{Credit}} & \multicolumn{4}{c|}{\textbf{Ionosphere}} \\
\cline{2-9}          & \textbf{FN} & \textbf{FP} & \textbf{CE} & \textbf{NEC} & \textbf{FN} & \textbf{FP} & \textbf{CE} & \textbf{NEC}  \\
    \hline
    \textbf{[1,100]} & 9.97$\cdot10^{-1}$ & 1.43$\cdot10^{-3}$ & 3.00$\cdot10^{-1}$ & 1.13$\cdot10^{-2}$ & 8.84$\cdot10^{-1}$ & 2.38$\cdot10^{-2}$ & 5.75$\cdot10^{-1}$ & 3.23$\cdot10^{-2}$ \\
    \textbf{[1,50]} & 9.97$\cdot10^{-1}$ & 1.43$\cdot10^{-3}$ & 3.00$\cdot10^{-1}$ & 2.09$\cdot10^{-2}$ & 8.93$\cdot10^{-1}$ & 1.59$\cdot10^{-2}$ & 5.78$\cdot10^{-1}$ & 3.31$\cdot10^{-2}$ \\
    \textbf{[1,25]} & 9.93$\cdot10^{-1}$ & 1.43$\cdot10^{-3}$ & 2.99$\cdot10^{-1}$ & 3.96$\cdot10^{-2}$ & 5.51$\cdot10^{-1}$ & 8.73$\cdot10^{-2}$ & 3.85$\cdot10^{-1}$ & 1.05$\cdot10^{-1}$ \\
    \textbf{[1,10]} & 9.40$\cdot10^{-1}$ & 5.72$\cdot10^{-3}$ & 2.86$\cdot10^{-1}$ & 9.07$\cdot10^{-2}$ & 4.44$\cdot10^{-1}$ & 8.73$\cdot10^{-2}$ & 3.16$\cdot10^{-1}$ & 1.20$\cdot10^{-1}$ \\
    \textbf{[1,7]} & 8.97$\cdot10^{-1}$ & 1.72$\cdot10^{-2}$ & 2.81$\cdot10^{-1}$ & 1.27$\cdot10^{-1}$ & 3.42$\cdot10^{-1}$ & 1.27$\cdot10^{-1}$ & 2.65$\cdot10^{-1}$ & 1.54$\cdot10^{-1}$ \\
    \textbf{[1,5]} & 8.43$\cdot10^{-1}$ & 3.00$\cdot10^{-2}$ & 2.74$\cdot10^{-1}$ & 1.66$\cdot10^{-1}$ & 2.67$\cdot10^{-1}$ & 1.51$\cdot10^{-1}$ & 2.25$\cdot10^{-1}$ & 1.70$\cdot10^{-1}$ \\
    \textbf{[1,3]} & 6.67$\cdot10^{-1}$ & 8.73$\cdot10^{-2}$ & 2.61$\cdot10^{-1}$ & 2.32$\cdot10^{-1}$ & 2.36$\cdot10^{-1}$ & 2.46$\cdot10^{-1}$ & 2.39$\cdot10^{-1}$ & 2.43$\cdot10^{-1}$ \\
    \textbf{[1,2]} & 5.03$\cdot10^{-1}$ & 1.23$\cdot10^{-1}$ & 2.37$\cdot10^{-1}$ & 2.50$\cdot10^{-1}$ & 1.29$\cdot10^{-1}$ & 2.46$\cdot10^{-1}$ & 1.71$\cdot10^{-1}$ & 2.07$\cdot10^{-1}$ \\
    \textbf{[2,3]} & 4.17$\cdot10^{-1}$ & 2.02$\cdot10^{-1}$ & 2.66$\cdot10^{-1}$ & 2.88$\cdot10^{-1}$ & 1.69$\cdot10^{-1}$ & 2.38$\cdot10^{-1}$ & 1.94$\cdot10^{-1}$ & 2.10$\cdot10^{-1}$ \\
    \textbf{[1,1]} & 2.60$\cdot10^{-1}$ & 2.90$\cdot10^{-1}$ & 2.81$\cdot10^{-1}$ & 2.75$\cdot10^{-1}$ & 3.56$\cdot10^{-2}$ & 3.02$\cdot10^{-1}$ & 1.31$\cdot10^{-1}$ & 1.69$\cdot10^{-1}$ \\
    \textbf{[3,2]} & 1.77$\cdot10^{-1}$ & 4.03$\cdot10^{-1}$ & 3.35$\cdot10^{-1}$ & 2.67$\cdot10^{-1}$ & 6.67$\cdot10^{-2}$ & 3.65$\cdot10^{-1}$ & 1.74$\cdot10^{-1}$ & 1.86$\cdot10^{-1}$ \\
    \textbf{[2,1]} & 1.47$\cdot10^{-1}$ & 4.38$\cdot10^{-1}$ & 3.50$\cdot10^{-1}$ & 2.44$\cdot10^{-1}$ & 5.33$\cdot10^{-2}$ & 3.17$\cdot10^{-1}$ & 1.48$\cdot10^{-1}$ & 1.41$\cdot10^{-1}$ \\
    \textbf{[3,1]} & 1.20$\cdot10^{-1}$ & 5.29$\cdot10^{-1}$ & 4.06$\cdot10^{-1}$ & 2.22$\cdot10^{-1}$ & 5.33$\cdot10^{-2}$ & 3.49$\cdot10^{-1}$ & 1.60$\cdot10^{-1}$ & 1.27$\cdot10^{-1}$ \\
    \textbf{[5,1]} & 7.33$\cdot10^{-2}$ & 6.74$\cdot10^{-1}$ & 4.93$\cdot10^{-1}$ & 1.73$\cdot10^{-1}$ & 3.56$\cdot10^{-2}$ & 3.10$\cdot10^{-1}$ & 1.34$\cdot10^{-1}$ & 8.12$\cdot10^{-2}$ \\
    \textbf{[7,1]} & 4.67$\cdot10^{-2}$ & 7.32$\cdot10^{-1}$ & 5.27$\cdot10^{-1}$ & 1.32$\cdot10^{-1}$ & 2.67$\cdot10^{-2}$ & 3.57$\cdot10^{-1}$ & 1.45$\cdot10^{-1}$ & 6.80$\cdot10^{-2}$ \\
    \textbf{[10,1]} & 2.33$\cdot10^{-2}$ & 8.18$\cdot10^{-1}$ & 5.80$\cdot10^{-1}$ & 9.56$\cdot10^{-2}$ & 3.56$\cdot10^{-2}$ & 3.73$\cdot10^{-1}$ & 1.57$\cdot10^{-1}$ & 6.62$\cdot10^{-2}$ \\
    \textbf{[25,1]} & 3.33$\cdot10^{-3}$ & 9.28$\cdot10^{-1}$ & 6.51$\cdot10^{-1}$ & 3.89$\cdot10^{-2}$ & 4.00$\cdot10^{-2}$ & 3.97$\cdot10^{-1}$ & 1.68$\cdot10^{-1}$ & 5.37$\cdot10^{-2}$ \\
    \textbf{[50,1]} & 0 & 9.67$\cdot10^{-1}$ & 6.77$\cdot10^{-1}$ & 1.90$\cdot10^{-2}$ & 4.44$\cdot10^{-2}$ & 4.05$\cdot10^{-1}$ & 1.74$\cdot10^{-1}$ & 5.15$\cdot10^{-2}$ \\
    \textbf{[100,1]} & 0 & 9.87$\cdot10^{-1}$ & 6.91$\cdot10^{-1}$ & 9.77$\cdot10^{-3}$ & 4.44$\cdot10^{-2}$ & 3.89$\cdot10^{-1}$ & 1.68$\cdot10^{-1}$ & 4.79$\cdot10^{-2}$ \\
    \hline
    \multicolumn{1}{c}{} & \multicolumn{1}{c}{} & \multicolumn{1}{c}{} & \multicolumn{1}{c}{} & \multicolumn{1}{c}{} & \multicolumn{1}{c}{} & \multicolumn{1}{c}{} & \multicolumn{1}{c}{} & \multicolumn{1}{c}{} \\
    \hline
    \multirow{2}[4]{*}{\textbf{Cost}} & \multicolumn{4}{c|}{\textbf{Diabetes}} & \multicolumn{4}{c|}{\textbf{Spam}} \\
\cline{2-9}          & \textbf{FN} & \textbf{FP} & \textbf{CE} & \textbf{NEC} & \textbf{FN} & \textbf{FP} & \textbf{CE} & \textbf{NEC}  \\
    \hline
    \textbf{[1,100]} & 9.81$\cdot10^{-1}$ & 4.02$\cdot10^{-3}$ & 3.45$\cdot10^{-1}$ & 1.37$\cdot10^{-2}$ & 3.89$\cdot10^{-1}$ & 1.38$\cdot10^{-2}$ & 2.41$\cdot10^{-1}$ & 1.75$\cdot10^{-2}$ \\
    \textbf{[1,50]} & 9.59$\cdot10^{-1}$ & 6.02$\cdot10^{-3}$ & 3.39$\cdot10^{-1}$ & 2.47$\cdot10^{-2}$ & 3.46$\cdot10^{-1}$ & 1.99$\cdot10^{-2}$ & 2.17$\cdot10^{-1}$ & 2.63$\cdot10^{-2}$ \\
    \textbf{[1,25]} & 9.03$\cdot10^{-1}$ & 1.61$\cdot10^{-2}$ & 3.25$\cdot10^{-1}$ & 5.02$\cdot10^{-2}$ & 2.74$\cdot10^{-1}$ & 2.76$\cdot10^{-2}$ & 1.77$\cdot10^{-1}$ & 3.71$\cdot10^{-2}$ \\
    \textbf{[1,10]} & 7.87$\cdot10^{-1}$ & 3.41$\cdot10^{-2}$ & 2.97$\cdot10^{-1}$ & 1.03$\cdot10^{-1}$ & 1.92$\cdot10^{-1}$ & 4.42$\cdot10^{-2}$ & 1.34$\cdot10^{-1}$ & 5.76$\cdot10^{-2}$ \\
    \textbf{[1,7]} & 6.52$\cdot10^{-1}$ & 4.02$\cdot10^{-2}$ & 2.54$\cdot10^{-1}$ & 1.17$\cdot10^{-1}$ & 1.69$\cdot10^{-1}$ & 4.75$\cdot10^{-2}$ & 1.21$\cdot10^{-1}$ & 6.27$\cdot10^{-2}$ \\
    \textbf{[1,5]} & 6.48$\cdot10^{-1}$ & 4.22$\cdot10^{-2}$ & 2.54$\cdot10^{-1}$ & 1.43$\cdot10^{-1}$ & 1.56$\cdot10^{-1}$ & 5.41$\cdot10^{-2}$ & 1.16$\cdot10^{-1}$ & 7.10$\cdot10^{-2}$ \\
    \textbf{[1,3]} & 5.62$\cdot10^{-1}$ & 6.63$\cdot10^{-2}$ & 2.39$\cdot10^{-1}$ & 1.90$\cdot10^{-1}$ & 1.23$\cdot10^{-1}$ & 6.46$\cdot10^{-2}$ & 9.98$\cdot10^{-2}$ & 7.91$\cdot10^{-2}$ \\
    \textbf{[1,2]} & 4.79$\cdot10^{-1}$ & 1.20$\cdot10^{-1}$ & 2.46$\cdot10^{-1}$ & 2.40$\cdot10^{-1}$ & 1.05$\cdot10^{-1}$ & 6.73$\cdot10^{-2}$ & 9.00$\cdot10^{-2}$ & 7.98$\cdot10^{-2}$ \\
    \textbf{[2,3]} & 3.56$\cdot10^{-1}$ & 1.99$\cdot10^{-1}$ & 2.54$\cdot10^{-1}$ & 2.62$\cdot10^{-1}$ & 8.93$\cdot10^{-2}$ & 6.84$\cdot10^{-2}$ & 8.11$\cdot10^{-2}$ & 7.68$\cdot10^{-2}$ \\
    \textbf{[1,1]} & 3.03$\cdot10^{-1}$ & 2.31$\cdot10^{-1}$ & 2.56$\cdot10^{-1}$ & 2.67$\cdot10^{-1}$ & 7.93$\cdot10^{-2}$ & 8.28$\cdot10^{-2}$ & 8.07$\cdot10^{-2}$ & 8.10$\cdot10^{-2}$ \\
    \textbf{[3,2]} & 2.40$\cdot10^{-1}$ & 3.03$\cdot10^{-1}$ & 2.81$\cdot10^{-1}$ & 2.65$\cdot10^{-1}$ & 7.18$\cdot10^{-2}$ & 9.11$\cdot10^{-2}$ & 7.94$\cdot10^{-2}$ & 7.95$\cdot10^{-2}$ \\
    \textbf{[2,1]} & 1.57$\cdot10^{-1}$ & 3.71$\cdot10^{-1}$ & 2.97$\cdot10^{-1}$ & 2.29$\cdot10^{-1}$ & 6.28$\cdot10^{-2}$ & 9.33$\cdot10^{-2}$ & 7.48$\cdot10^{-2}$ & 7.30$\cdot10^{-2}$ \\
    \textbf{[3,1]} & 1.42$\cdot10^{-1}$ & 4.32$\cdot10^{-1}$ & 3.31$\cdot10^{-1}$ & 2.15$\cdot10^{-1}$ & 5.81$\cdot10^{-2}$ & 1.13$\cdot10^{-1}$ & 7.98$\cdot10^{-2}$ & 7.19$\cdot10^{-2}$ \\
    \textbf{[5,1]} & 9.36$\cdot10^{-2}$ & 5.16$\cdot10^{-1}$ & 3.69$\cdot10^{-1}$ & 1.64$\cdot10^{-1}$ & 4.99$\cdot10^{-2}$ & 1.41$\cdot10^{-1}$ & 8.57$\cdot10^{-2}$ & 6.50$\cdot10^{-2}$ \\
    \textbf{[7,1]} & 8.99$\cdot10^{-2}$ & 5.42$\cdot10^{-1}$ & 3.84$\cdot10^{-1}$ & 1.46$\cdot10^{-1}$ & 4.49$\cdot10^{-2}$ & 1.46$\cdot10^{-1}$ & 8.46$\cdot10^{-2}$ & 5.75$\cdot10^{-2}$ \\
    \textbf{[10,1]} & 7.12$\cdot10^{-2}$ & 5.94$\cdot10^{-1}$ & 4.12$\cdot10^{-1}$ & 1.19$\cdot10^{-1}$ & 4.31$\cdot10^{-2}$ & 1.69$\cdot10^{-1}$ & 9.26$\cdot10^{-2}$ & 5.45$\cdot10^{-2}$ \\
    \textbf{[25,1]} & 4.49$\cdot10^{-2}$ & 6.79$\cdot10^{-1}$ & 4.58$\cdot10^{-1}$ & 6.93$\cdot10^{-2}$ & 3.98$\cdot10^{-2}$ & 2.16$\cdot10^{-1}$ & 1.09$\cdot10^{-1}$ & 4.66$\cdot10^{-2}$ \\
    \textbf{[50,1]} & 2.62$\cdot10^{-2}$ & 7.27$\cdot10^{-1}$ & 4.82$\cdot10^{-1}$ & 4.00$\cdot10^{-2}$ & 3.41$\cdot10^{-2}$ & 2.41$\cdot10^{-1}$ & 1.15$\cdot10^{-1}$ & 3.81$\cdot10^{-2}$ \\
    \textbf{[100,1]} & 1.50$\cdot10^{-2}$ & 7.65$\cdot10^{-1}$ & 5.03$\cdot10^{-1}$ & 2.24$\cdot10^{-2}$ & 2.94$\cdot10^{-2}$ & 2.81$\cdot10^{-1}$ & 1.29$\cdot10^{-1}$ & 3.19$\cdot10^{-2}$ \\
    \hline
    \end{tabular}}%
  \label{asym_perform_tab}%
\end{table*}%

\subsection{AdaBoostDB vs. Cost-Sensitive AdaBoost}
\label{subsec:ADB_vs_CSB}

As explained in section \ref{sec:AdaBoostDB}, AdaBoostDB and Cost-Sensitive AdaBoost share a common theoretical root, but differ in the way they model and derive that equivalent starting point. As a result, both frameworks give rise to different algorithms that must obtain the same solution for a given problem. This scenario has two consequences: on the one side, though classifiers obtained by both algorithms should be theoretically identical when trained in the same conditions, in practice numerical errors can make them differ. On the other side, the polynomial model and Conditional Search mechanism related to AdaBoostDB entails differences in computing time which should be quantified. In this section we will comparatively evaluate these aspects.

It is important to highlight that Cost-Sensitive AdaBoost has been shown to outperform other previous asymmetric approaches in the literature, as can be seen in \cite{MasnadiVasconcelos11} and \cite{MasnadiVasconcelos07}. As a consequence, in order to avoid redundant experiments already published in other works, we have focused our efforts on comparing our method with Cost-Sensitive AdaBoost and demonstrate that, thought very different in computational burden, both algorithms are equivalent in classification performance. We encourage the reader to consult \cite{MasnadiVasconcelos11} and \cite{MasnadiVasconcelos07} to deepen the comparison with other algorithms, since, as we will see, classification performance differences between Cost-Sensitive AdaBoost and AdaBoostDB, only due to numerical errors, are negligible.

\subsubsection{Classification Performance}
\label{perform_cmp}

As we have just commented, though theoretically equivalent, classifiers obtained from AdaBoostDB and Cost-Sensitive AdaBoost tend to differ due to numerical errors related to the different model (polynomial vs. hyperbolic) adopted in each case. In section \ref{sub_sec:bayes_risk} we have seen that differences in the Bayes scenario are negligible. To further test the relevance of this difference, we have used again the same datasets, cost combinations and 3-fold cross-validation strategy used in the last section, now applied to Cost-Sensitive AdaBoost. 

\begin{figure}[ht]
\begin{center}
\centerline{\includegraphics[width=3.5in]{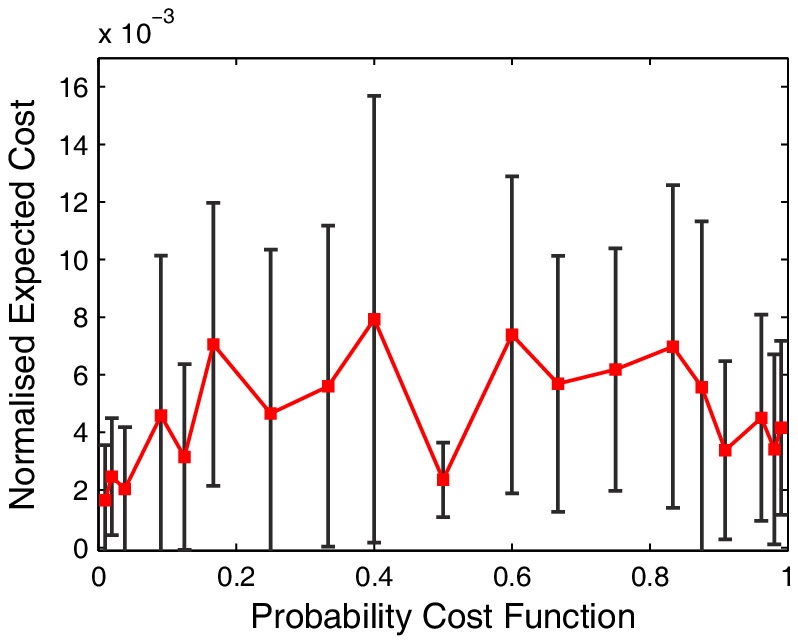}}
\caption{Mean Error and Standard Deviation of the Normalized Expected Cost between AdaBoostDB and Cost-Sensitive Boosting across all the datasets and for every cost-combination.}
\label{db_vs_csb}
\end{center}
\end{figure}

The mean error between the two alternatives is tabulated in Table \ref{diffs_tab}, and, as can be seen, is only the order of hundredths for the worst case. To make a more visual interpretation of this differences, we have also computed the mean and standard deviation across all the datasets, of the Normalized Expected Cost (the more accurate single measure of asymmetric performance) for every trained cost-combination. The result can be seen in Figure \ref{db_vs_csb}, where differences are in the range of thousandths. As we could expect, classification performance differences are again negligible in all cases.

\begin{table*}[htbp]
  \centering
  \footnotesize
  \caption{Mean error between AdaBoostDB and Cost-Sensitive Boosting. It has been computed across the 3 cross-validation training and test sets of every dataset and all trained rounds.}
  \scalebox{0.6}{    
    \begin{tabular}{|c|c|c|c|c|c|c|c|c|}
    \hline
    \multirow{2}[4]{*}{\textbf{Cost}} & \multicolumn{4}{c|}{\textbf{Bayes}} & \multicolumn{4}{c|}{\textbf{TwoClouds}} \\
\cline{2-9}          & \textbf{FN} & \textbf{FP} & \textbf{CE} & \textbf{NEC} & \textbf{FN} & \textbf{FP} & \textbf{CE} & \textbf{NEC} \\
    \hline
    \textbf{[1,100]} & 6.48$\cdot10^{-2}$ & 6.02$\cdot10^{-3}$ & 2.99$\cdot10^{-2}$ & 5.65$\cdot10^{-3}$ & 0     & 0     & 0     & 0 \\
    \textbf{[1,50]} & 7.23$\cdot10^{-2}$ & 8.53$\cdot10^{-3}$ & 3.34$\cdot10^{-2}$ & 7.54$\cdot10^{-3}$ & 4.82$\cdot10^{-2}$ & 4.02$\cdot10^{-3}$ & 2.21$\cdot10^{-2}$ & 2.99$\cdot10^{-3}$ \\
    \textbf{[1,25]} & 2.56$\cdot10^{-2}$ & 5.52$\cdot10^{-3}$ & 1.51$\cdot10^{-2}$ & 6.02$\cdot10^{-3}$ & 0     & 0     & 0     & 0 \\
    \textbf{[1,10]} & 3.46$\cdot10^{-2}$ & 6.53$\cdot10^{-3}$ & 1.61$\cdot10^{-2}$ & 4.15$\cdot10^{-3}$ & 8.39$\cdot10^{-3}$ & 4.13$\cdot10^{-3}$ & 2.13$\cdot10^{-3}$ & 3.00$\cdot10^{-3}$ \\
    \textbf{[1,7]} & 4.82$\cdot10^{-2}$ & 5.02$\cdot10^{-3}$ & 2.16$\cdot10^{-2}$ & 3.14$\cdot10^{-3}$ & 5.82$\cdot10^{-2}$ & 7.32$\cdot10^{-3}$ & 2.59$\cdot10^{-2}$ & 2.47$\cdot10^{-3}$ \\
    \textbf{[1,5]} & 4.27$\cdot10^{-2}$ & 1.05$\cdot10^{-2}$ & 1.66$\cdot10^{-2}$ & 6.86$\cdot10^{-3}$ & 3.43$\cdot10^{-2}$ & 1.57$\cdot10^{-2}$ & 2.16$\cdot10^{-2}$ & 1.64$\cdot10^{-2}$ \\
    \textbf{[1,3]} & 2.76$\cdot10^{-2}$ & 6.53$\cdot10^{-3}$ & 1.05$\cdot10^{-2}$ & 4.52$\cdot10^{-3}$ & 1.91$\cdot10^{-2}$ & 2.83$\cdot10^{-3}$ & 8.39$\cdot10^{-3}$ & 3.31$\cdot10^{-3}$ \\
    \textbf{[1,2]} & 2.81$\cdot10^{-2}$ & 1.46$\cdot10^{-2}$ & 6.78$\cdot10^{-3}$ & 3.35$\cdot10^{-4}$ & 2.35$\cdot10^{-2}$ & 8.15$\cdot10^{-3}$ & 7.91$\cdot10^{-3}$ & 3.27$\cdot10^{-3}$ \\
    \textbf{[2,3]} & 4.02$\cdot10^{-3}$ & 0     & 2.01$\cdot10^{-3}$ & 1.61$\cdot10^{-3}$ & 4.65$\cdot10^{-2}$ & 3.69$\cdot10^{-2}$ & 1.35$\cdot10^{-2}$ & 1.41$\cdot10^{-2}$ \\
    \textbf{[1,1]} & 0     & 0     & 0     & 0     & 6.67$\cdot10^{-2}$ & 7.54$\cdot10^{-2}$ & 1.41$\cdot10^{-2}$ & 1.41$\cdot10^{-2}$ \\
    \textbf{[3,2]} & 3.01$\cdot10^{-3}$ & 1.51$\cdot10^{-2}$ & 7.53$\cdot10^{-3}$ & 6.02$\cdot10^{-3}$ & 7.22$\cdot10^{-2}$ & 6.11$\cdot10^{-2}$ & 1.69$\cdot10^{-2}$ & 2.37$\cdot10^{-2}$ \\
    \textbf{[2,1]} & 1.41$\cdot10^{-2}$ & 1.26$\cdot10^{-2}$ & 1.26$\cdot10^{-3}$ & 5.19$\cdot10^{-3}$ & 3.54$\cdot10^{-3}$ & 4.61$\cdot10^{-3}$ & 1.24$\cdot10^{-3}$ & 1.30$\cdot10^{-3}$ \\
    \textbf{[3,1]} & 9.54$\cdot10^{-3}$ & 3.82$\cdot10^{-2}$ & 1.43$\cdot10^{-2}$ & 6.65$\cdot10^{-3}$ & 1.65$\cdot10^{-3}$ & 5.08$\cdot10^{-3}$ & 1.83$\cdot10^{-3}$ & 9.74$\cdot10^{-4}$ \\
    \textbf{[5,1]} & 1.10$\cdot10^{-2}$ & 4.82$\cdot10^{-2}$ & 1.91$\cdot10^{-2}$ & 3.85$\cdot10^{-3}$ & 9.45$\cdot10^{-3}$ & 2.48$\cdot10^{-2}$ & 1.38$\cdot10^{-2}$ & 8.43$\cdot10^{-3}$ \\
    \textbf{[7,1]} & 1.36$\cdot10^{-2}$ & 5.62$\cdot10^{-2}$ & 2.13$\cdot10^{-2}$ & 5.46$\cdot10^{-3}$ & 1.65$\cdot10^{-3}$ & 1.18$\cdot10^{-2}$ & 6.73$\cdot10^{-3}$ & 2.92$\cdot10^{-3}$ \\
    \textbf{[10,1]} & 6.53$\cdot10^{-3}$ & 2.31$\cdot10^{-2}$ & 1.08$\cdot10^{-2}$ & 5.84$\cdot10^{-3}$ & 0 & 5.20$\cdot10^{-3}$ & 2.60$\cdot10^{-3}$ & 4.72$\cdot10^{-4}$ \\
    \textbf{[25,1]} & 6.53$\cdot10^{-3}$ & 3.41$\cdot10^{-2}$ & 1.38$\cdot10^{-2}$ & 5.12$\cdot10^{-3}$ & 4.02$\cdot10^{-3}$ & 2.01$\cdot10^{-3}$ & 1.00$\cdot10^{-3}$ & 3.78$\cdot10^{-3}$ \\
    \textbf{[50,1]} & 8.53$\cdot10^{-3}$ & 4.37$\cdot10^{-2}$ & 1.76$\cdot10^{-2}$ & 7.51$\cdot10^{-3}$ & 0     & 0     & 0     & 0 \\
    \textbf{[100,1]} & 9.04$\cdot10^{-3}$ & 4.77$\cdot10^{-2}$ & 1.93$\cdot10^{-2}$ & 8.47$\cdot10^{-3}$ & 4.02$\cdot10^{-3}$ & 2.01$\cdot10^{-3}$ & 1.00$\cdot10^{-3}$ & 3.96$\cdot10^{-3}$ \\
    \hline
    \multicolumn{1}{c}{} & \multicolumn{1}{c}{} & \multicolumn{1}{c}{} & \multicolumn{1}{c}{} & \multicolumn{1}{c}{} & \multicolumn{1}{c}{} & \multicolumn{1}{c}{} & \multicolumn{1}{c}{} & \multicolumn{1}{c}{} \\
    \hline
    \multirow{2}[4]{*}{\textbf{Cost}} & \multicolumn{4}{c|}{\textbf{Credit}} & \multicolumn{4}{c|}{\textbf{Ionosphere}} \\
\cline{2-9}          & \textbf{FN} & \textbf{FP} & \textbf{CE} & \textbf{NEC} & \textbf{FN} & \textbf{FP} & \textbf{CE} & \textbf{NEC}\\
    \hline
    \textbf{[1,100]} & 0     & 0     & 0     & 0     & 4.44$\cdot10^{-3}$ & 1.32$\cdot10^{-3}$ & 3.32$\cdot10^{-3}$ & 1.35$\cdot10^{-3}$ \\
    \textbf{[1,50]} & 0     & 0     & 0     & 0     & 3.04$\cdot10^{-2}$ & 1.32$\cdot10^{-3}$ & 1.90$\cdot10^{-2}$ & 9.34$\cdot10^{-4}$ \\
    \textbf{[1,25]} & 3.92$\cdot10^{-4}$ & 0     & 1.18$\cdot10^{-4}$ & 1.51$\cdot10^{-5}$ & 3.48$\cdot10^{-2}$ & 2.65$\cdot10^{-3}$ & 2.14$\cdot10^{-2}$ & 1.20$\cdot10^{-3}$ \\
    \textbf{[1,10]} & 4.12$\cdot10^{-3}$ & 1.01$\cdot10^{-3}$ & 1.59$\cdot10^{-3}$ & 1.15$\cdot10^{-3}$ & 1.78$\cdot10^{-2}$ & 1.19$\cdot10^{-2}$ & 9.02$\cdot10^{-3}$ & 9.75$\cdot10^{-3}$ \\
    \textbf{[1,7]} & 1.12$\cdot10^{-2}$ & 2.95$\cdot10^{-3}$ & 2.12$\cdot10^{-3}$ & 1.96$\cdot10^{-3}$ & 8.89$\cdot10^{-3}$ & 5.29$\cdot10^{-3}$ & 3.80$\cdot10^{-3}$ & 3.52$\cdot10^{-3}$ \\
    \textbf{[1,5]} & 2.67$\cdot10^{-2}$ & 8.42$\cdot10^{-3}$ & 5.06$\cdot10^{-3}$ & 4.66$\cdot10^{-3}$ & 4.30$\cdot10^{-2}$ & 2.65$\cdot10^{-3}$ & 2.66$\cdot10^{-2}$ & 4.96$\cdot10^{-3}$ \\
    \textbf{[1,3]} & 2.12$\cdot10^{-2}$ & 8.92$\cdot10^{-3}$ & 2.24$\cdot10^{-3}$ & 3.22$\cdot10^{-3}$ & 1.85$\cdot10^{-2}$ & 1.59$\cdot10^{-2}$ & 1.38$\cdot10^{-2}$ & 1.13$\cdot10^{-2}$ \\
    \textbf{[1,2]} & 2.33$\cdot10^{-2}$ & 1.48$\cdot10^{-2}$ & 5.95$\cdot10^{-3}$ & 5.31$\cdot10^{-3}$ & 8.37$\cdot10^{-2}$ & 3.70$\cdot10^{-2}$ & 4.51$\cdot10^{-2}$ & 1.62$\cdot10^{-2}$ \\
    \textbf{[2,3]} & 5.49$\cdot10^{-3}$ & 2.36$\cdot10^{-3}$ & 1.18$\cdot10^{-3}$ & 1.62$\cdot10^{-3}$ & 2.59$\cdot10^{-2}$ & 2.38$\cdot10^{-2}$ & 1.85$\cdot10^{-2}$ & 2.05$\cdot10^{-2}$ \\
    \textbf{[1,1]} & 0     & 0     & 0     & 0     & 0     & 0     & 0     & 0 \\
    \textbf{[3,2]} & 5.88$\cdot10^{-3}$ & 7.15$\cdot10^{-3}$ & 4.53$\cdot10^{-3}$ & 3.87$\cdot10^{-3}$ & 1.41$\cdot10^{-2}$ & 2.51$\cdot10^{-2}$ & 8.55$\cdot10^{-3}$ & 9.27$\cdot10^{-3}$ \\
    \textbf{[2,1]} & 7.65$\cdot10^{-3}$ & 9.43$\cdot10^{-3}$ & 4.89$\cdot10^{-3}$ & 2.70$\cdot10^{-3}$ & 5.93$\cdot10^{-3}$ & 2.25$\cdot10^{-2}$ & 9.97$\cdot10^{-3}$ & 9.47$\cdot10^{-3}$ \\
    \textbf{[3,1]} & 9.02$\cdot10^{-3}$ & 2.23$\cdot10^{-2}$ & 1.51$\cdot10^{-2}$ & 5.87$\cdot10^{-3}$ & 5.93$\cdot10^{-3}$ & 1.06$\cdot10^{-2}$ & 6.65$\cdot10^{-3}$ & 5.98$\cdot10^{-3}$ \\
    \textbf{[5,1]} & 4.51$\cdot10^{-3}$ & 1.00$\cdot10^{-2}$ & 6.24$\cdot10^{-3}$ & 3.31$\cdot10^{-3}$ & 8.89$\cdot10^{-3}$ & 2.51$\cdot10^{-2}$ & 1.28$\cdot10^{-2}$ & 1.07$\cdot10^{-2}$ \\
    \textbf{[7,1]} & 3.33$\cdot10^{-3}$ & 8.42$\cdot10^{-3}$ & 6.30$\cdot10^{-3}$ & 3.19$\cdot10^{-3}$ & 1.48$\cdot10^{-2}$ & 2.12$\cdot10^{-2}$ & 6.65$\cdot10^{-3}$ & 1.20$\cdot10^{-2}$ \\
    \textbf{[10,1]} & 0 & 1.26$\cdot10^{-3}$ & 8.83$\cdot10^{-4}$ & 1.15$\cdot10^{-4}$ & 6.67$\cdot10^{-3}$ & 1.72$\cdot10^{-2}$ & 2.85$\cdot10^{-3}$ & 4.74$\cdot10^{-3}$ \\
    \textbf{[25,1]} & 3.14$\cdot10^{-3}$ & 5.39$\cdot10^{-3}$ & 4.36$\cdot10^{-3}$ & 3.16$\cdot10^{-3}$ & 8.89$\cdot10^{-3}$ & 1.19$\cdot10^{-2}$ & 8.07$\cdot10^{-3}$ & 8.80$\cdot10^{-3}$ \\
    \textbf{[50,1]} & 0     & 3.79$\cdot10^{-3}$ & 2.65$\cdot10^{-3}$ & 7.43$\cdot10^{-5}$ & 9.63$\cdot10^{-3}$ & 1.85$\cdot10^{-2}$ & 6.17$\cdot10^{-3}$ & 9.13$\cdot10^{-3}$ \\
    \textbf{[100,1]} & 0     & 1.68$\cdot10^{-4}$ & 1.18$\cdot10^{-4}$ & 1.67$\cdot10^{-6}$ & 9.63$\cdot10^{-3}$ & 1.85$\cdot10^{-2}$ & 7.12$\cdot10^{-3}$ & 9.40$\cdot10^{-3}$ \\
    \hline
    \multicolumn{1}{c}{} & \multicolumn{1}{c}{} & \multicolumn{1}{c}{} & \multicolumn{1}{c}{} & \multicolumn{1}{c}{} & \multicolumn{1}{c}{} & \multicolumn{1}{c}{} & \multicolumn{1}{c}{} & \multicolumn{1}{c}{} \\
    \hline
    \multirow{2}[4]{*}{\textbf{Cost}} & \multicolumn{4}{c|}{\textbf{Diabetes}} & \multicolumn{4}{c|}{\textbf{Spam}} \\
\cline{2-9}          & \textbf{FN} & \textbf{FP} & \textbf{CE} & \textbf{NEC} & \textbf{FN} & \textbf{FP} & \textbf{CE} & \textbf{NEC}\\
    \hline
    \textbf{[1,100]} & 7.49$\cdot10^{-3}$ & 0     & 2.61$\cdot10^{-3}$ & 7.42$\cdot10^{-5}$ & 2.61$\cdot10^{-2}$ & 2.96$\cdot10^{-3}$ & 1.48$\cdot10^{-2}$ & 2.77$\cdot10^{-3}$ \\
    \textbf{[1,50]} & 8.93$\cdot10^{-3}$ & 0     & 3.12$\cdot10^{-3}$ & 1.75$\cdot10^{-4}$ & 2.30$\cdot10^{-2}$ & 3.43$\cdot10^{-3}$ & 1.33$\cdot10^{-2}$ & 3.15$\cdot10^{-3}$ \\
    \textbf{[1,25]} & 1.15$\cdot10^{-2}$ & 2.47$\cdot10^{-3}$ & 2.82$\cdot10^{-3}$ & 2.20$\cdot10^{-3}$ & 2.38$\cdot10^{-2}$ & 3.19$\cdot10^{-3}$ & 1.39$\cdot10^{-2}$ & 2.78$\cdot10^{-3}$ \\
    \textbf{[1,10]} & 1.87$\cdot10^{-2}$ & 6.33$\cdot10^{-3}$ & 7.24$\cdot10^{-3}$ & 5.78$\cdot10^{-3}$ & 1.47$\cdot10^{-2}$ & 4.52$\cdot10^{-3}$ & 8.28$\cdot10^{-3}$ & 3.70$\cdot10^{-3}$ \\
    \textbf{[1,7]} & 4.35$\cdot10^{-2}$ & 1.08$\cdot10^{-2}$ & 8.75$\cdot10^{-3}$ & 4.71$\cdot10^{-3}$ & 1.42$\cdot10^{-2}$ & 4.58$\cdot10^{-3}$ & 7.91$\cdot10^{-3}$ & 3.11$\cdot10^{-3}$ \\
    \textbf{[1,5]} & 4.64$\cdot10^{-2}$ & 6.02$\cdot10^{-3}$ & 1.29$\cdot10^{-2}$ & 5.47$\cdot10^{-3}$ & 1.64$\cdot10^{-2}$ & 5.18$\cdot10^{-3}$ & 9.06$\cdot10^{-3}$ & 4.01$\cdot10^{-3}$ \\
    \textbf{[1,3]} & 4.61$\cdot10^{-3}$ & 9.27$\cdot10^{-4}$ & 1.61$\cdot10^{-3}$ & 1.15$\cdot10^{-3}$ & 1.29$\cdot10^{-2}$ & 7.07$\cdot10^{-3}$ & 6.70$\cdot10^{-3}$ & 4.41$\cdot10^{-3}$ \\
    \textbf{[1,2]} & 8.93$\cdot10^{-3}$ & 6.02$\cdot10^{-3}$ & 6.03$\cdot10^{-3}$ & 5.99$\cdot10^{-3}$ & 9.44$\cdot10^{-3}$ & 3.17$\cdot10^{-3}$ & 5.16$\cdot10^{-3}$ & 2.54$\cdot10^{-3}$ \\
    \textbf{[2,3]} & 1.61$\cdot10^{-2}$ & 9.42$\cdot10^{-3}$ & 5.93$\cdot10^{-3}$ & 6.24$\cdot10^{-3}$ & 8.19$\cdot10^{-3}$ & 3.44$\cdot10^{-3}$ & 4.98$\cdot10^{-3}$ & 3.54$\cdot10^{-3}$ \\
    \textbf{[1,1]} & 0     & 0     & 0     & 0     & 0     & 0     & 0     & 0 \\
    \textbf{[3,2]} & 1.15$\cdot10^{-3}$ & 1.70$\cdot10^{-3}$ & 7.04$\cdot10^{-4}$ & 6.30$\cdot10^{-4}$ & 1.08$\cdot10^{-3}$ & 2.64$\cdot10^{-3}$ & 7.79$\cdot10^{-4}$ & 7.90$\cdot10^{-4}$ \\
    \textbf{[2,1]} & 3.23$\cdot10^{-2}$ & 2.97$\cdot10^{-2}$ & 9.05$\cdot10^{-3}$ & 1.23$\cdot10^{-2}$ & 5.75$\cdot10^{-3}$ & 5.95$\cdot10^{-3}$ & 2.90$\cdot10^{-3}$ & 3.12$\cdot10^{-3}$ \\
    \textbf{[3,1]} & 2.30$\cdot10^{-2}$ & 3.82$\cdot10^{-2}$ & 1.90$\cdot10^{-2}$ & 1.38$\cdot10^{-2}$ & 7.97$\cdot10^{-3}$ & 1.48$\cdot10^{-2}$ & 4.24$\cdot10^{-3}$ & 3.83$\cdot10^{-3}$ \\
    \textbf{[5,1]} & 1.61$\cdot10^{-2}$ & 2.73$\cdot10^{-2}$ & 1.38$\cdot10^{-2}$ & 9.40$\cdot10^{-3}$ & 6.89$\cdot10^{-3}$ & 8.06$\cdot10^{-3}$ & 5.40$\cdot10^{-3}$ & 6.20$\cdot10^{-3}$ \\
    \textbf{[7,1]} & 1.07$\cdot10^{-2}$ & 3.49$\cdot10^{-2}$ & 1.98$\cdot10^{-2}$ & 5.93$\cdot10^{-3}$ & 4.63$\cdot10^{-3}$ & 1.43$\cdot10^{-2}$ & 5.68$\cdot10^{-3}$ & 3.98$\cdot10^{-3}$ \\
    \textbf{[10,1]} & 8.93$\cdot10^{-3}$ & 3.88$\cdot10^{-2}$ & 2.43$\cdot10^{-2}$ & 6.30$\cdot10^{-3}$ & 3.46$\cdot10^{-3}$ & 1.71$\cdot10^{-2}$ & 6.05$\cdot10^{-3}$ & 2.82$\cdot10^{-3}$ \\
    \textbf{[25,1]} & 4.32$\cdot10^{-3}$ & 2.87$\cdot10^{-2}$ & 1.72$\cdot10^{-2}$ & 3.13$\cdot10^{-3}$ & 3.38$\cdot10^{-3}$ & 1.57$\cdot10^{-2}$ & 5.81$\cdot10^{-3}$ & 3.07$\cdot10^{-3}$ \\
    \textbf{[50,1]} & 1.73$\cdot10^{-3}$ & 2.80$\cdot10^{-2}$ & 1.82$\cdot10^{-2}$ & 1.87$\cdot10^{-3}$ & 1.89$\cdot10^{-3}$ & 1.38$\cdot10^{-2}$ & 5.67$\cdot10^{-3}$ & 1.90$\cdot10^{-3}$ \\
    \textbf{[100,1]} & 5.76$\cdot10^{-4}$ & 2.32$\cdot10^{-2}$ & 1.49$\cdot10^{-2}$ & 6.90$\cdot10^{-4}$ & 2.47$\cdot10^{-3}$ & 1.90$\cdot10^{-2}$ & 7.62$\cdot10^{-3}$ & 2.45$\cdot10^{-3}$ \\
    \hline
    \end{tabular}}%
  \label{diffs_tab}%
\end{table*}%

\subsubsection{Computation Time}
\label{sec:time_cmp}

The next item of our empirical comparison is quantifying, in terms of time and number of evaluated zeros, the accelerating power of AdaBoostDB respect to Cost-Sensitive Boosting. For this task, we have recorded the time consumed to train all the classifiers used in the previous tests for AdaBoostDB and Cost-Sensitive AdaBoost, plus one more variation: AdaBoostDB is also computed without the Conditional Search, in order to evaluate how much time saving would be attributable only to the polynomial model, leaving apart the Conditional Search.

\begin{table*}[htbp]
    \centering
  \footnotesize
  \caption{Training computational burden (number of zero searches and elapsed time in seconds) of Cost-Sensitive AdaBoost [CS], AdaBoostDB without conditional search [DBN], and AdaBoostDB (with conditional search) [DB] over the synthetic and UCI sets.}
  \scalebox{0.55}{
    \begin{tabular}{|c|r|c|c|c|c|c|c|c|c|c|c|c|c|}
    \hline
    \multirow{2}[4]{*}{\textbf{Cost}} & \multicolumn{1}{c|}{\multirow{2}[4]{*}{\textbf{Method}}} & \multicolumn{2}{c|}{\textbf{Bayes}} & \multicolumn{2}{c|}{\textbf{Two Clouds}} & \multicolumn{2}{c|}{\textbf{Credit}} & \multicolumn{2}{c|}{\textbf{Ionosphere}} & \multicolumn{2}{c|}{\textbf{Diabetes}} & \multicolumn{2}{c|}{\textbf{Spam}} \\
\cline{3-14}          & \multicolumn{1}{c|}{} & \textbf{Zeros} & \textbf{Time} & \textbf{Zeros} & \textbf{Time} & \textbf{Zeros} & \textbf{Time} & \textbf{Zeros} & \textbf{Time} & \textbf{Zeros} & \textbf{Time} & \textbf{Zeros} & \textbf{Time} \\
    \hline
    \multirow{3}[2]{*}{\textbf{[1,100]}} & \multicolumn{1}{c|}{\textbf{CS}} & 864528 & 579.76 & 3416944 & 2204.93 & 42607 & 27.52 & 317550 & 206.82 & 132992 & 84.76 & 9637557 & 6123.66 \\
          & \multicolumn{1}{c|}{\textbf{DBN}} & 864528 & 462.20 & 3416944 & 1606.21 & 42607 & 19.61 & 317550 & 154.51 & 132992 & 62.02 & 9637557 & 4547.29 \\
          & \multicolumn{1}{c|}{\textbf{DB}} & \textbf{3244} & \textbf{1.96} & \textbf{3575} & \textbf{3.86} & \textbf{606} & \textbf{0.43} & \textbf{698} & \textbf{0.59} & \textbf{647} & \textbf{0.44} & \textbf{23448} & \textbf{20.50} \\
    \hline
    \multirow{3}[2]{*}{\textbf{[1,50]}} & \multicolumn{1}{c|}{\textbf{CS}} & 864528 & 583.36 & 3416944 & 2183.68 & 42607 & 27.28 & 317550 & 206.94 & 132992 & 85.49 & 9637557 & 6012.60 \\
          & \multicolumn{1}{c|}{\textbf{DBN}} & 864528 & 450.55 & 3416944 & 1570.13 & 42607 & 19.76 & 317550 & 149.75 & 132992 & 60.79 & 9637557 & 4305.86 \\
          & \multicolumn{1}{c|}{\textbf{DB}} & \textbf{3626} & \textbf{2.27} & \textbf{3601} & \textbf{3.97} & \textbf{598} & \textbf{0.45} & \textbf{734} & \textbf{0.64} & \textbf{667} & \textbf{0.47} & \textbf{21641} & \textbf{19.65} \\
    \hline
    \multirow{3}[2]{*}{\textbf{[1,25]}} & \multicolumn{1}{c|}{\textbf{CS}} & 864528 & 570.09 & 3416944 & 2219.51 & 42607 & 27.02 & 317550 & 206.49 & 132992 & 89.05 & 9637557 & 6001.78 \\
          & \multicolumn{1}{c|}{\textbf{DBN}} & 864528 & 452.93 & 3416944 & 1537.65 & 42607 & 19.26 & 317550 & 151.34 & 132992 & 62.18 & 9637557 & 4226.14 \\
          & \multicolumn{1}{c|}{\textbf{DB}} & \textbf{3900} & \textbf{2.69} & \textbf{3601} & \textbf{3.97} & \textbf{635} & \textbf{0.46} & \textbf{660} & \textbf{0.55} & \textbf{655} & \textbf{0.45} & \textbf{19305} & \textbf{18.37} \\
    \hline
    \multirow{3}[2]{*}{\textbf{[1,10]}} & \multicolumn{1}{c|}{\textbf{CS}} & 864528 & 571.30 & 3416944 & 2181.71 & 42607 & 26.99 & 317550 & 206.37 & 132992 & 84.97 & 9637557 & 5983.12 \\
          & \multicolumn{1}{c|}{\textbf{DBN}} & 864528 & 462.02 & 3416944 & 1510.47 & 42607 & 18.68 & 317550 & 161.55 & 132992 & 58.62 & 9637557 & 4122.24 \\
          & \multicolumn{1}{c|}{\textbf{DB}} & \textbf{4496} & \textbf{2.91} & \textbf{3445} & \textbf{3.79} & \textbf{623} & \textbf{0.44} & \textbf{579} & \textbf{0.53} & \textbf{664} & \textbf{0.44} & \textbf{15804} & \textbf{16.21} \\
    \hline
    \multirow{3}[2]{*}{\textbf{[1,7]}} & \multicolumn{1}{c|}{\textbf{CS}} & 864528 & 563.62 & 3416944 & 2196.89 & 42607 & 27.39 & 317550 & 208.32 & 132992 & 84.82 & 9637557 & 5980.80 \\
          & \multicolumn{1}{c|}{\textbf{DBN}} & 864528 & 459.41 & 3416944 & 1531.74 & 42607 & 18.64 & 317550 & 166.99 & 132992 & 59.19 & 9637557 & 4124.47 \\
          & \multicolumn{1}{c|}{\textbf{DB}} & \textbf{4465} & \textbf{2.78} & \textbf{3406} & \textbf{3.75} & \textbf{632} & \textbf{0.42} & \textbf{596} & \textbf{0.50} & \textbf{687} & \textbf{0.46} & \textbf{14389} & \textbf{15.30} \\
    \hline
    \multirow{3}[2]{*}{\textbf{[1,5]}} & \multicolumn{1}{c|}{\textbf{CS}} & 864528 & 560.89 & 3416944 & 2122.42 & 42607 & 28.95 & 317550 & 208.91 & 132992 & 84.55 & 9637557 & 5955.99 \\
          & \multicolumn{1}{c|}{\textbf{DBN}} & 864528 & 467.03 & 3416944 & 1544.07 & 42607 & 20.48 & 317550 & 171.58 & 132992 & 59.41 & 9637557 & 4147.62 \\
          & \multicolumn{1}{c|}{\textbf{DB}} & \textbf{4403} & \textbf{2.69} & \textbf{3360} & \textbf{3.60} & \textbf{616} & \textbf{0.47} & \textbf{544} & \textbf{0.46} & \textbf{661} & \textbf{0.42} & \textbf{13122} & \textbf{14.58} \\
    \hline
    \multirow{3}[2]{*}{\textbf{[1,3]}} & \multicolumn{1}{c|}{\textbf{CS}} & 864528 & 559.56 & 3416944 & 2106.83 & 42607 & 28.49 & 317550 & 205.09 & 132992 & 84.61 & 9637557 & 5930.07 \\
          & \multicolumn{1}{c|}{\textbf{DBN}} & 864528 & 468.71 & 3416944 & 1555.67 & 42607 & 19.86 & 317550 & 174.67 & 132992 & 60.61 & 9637557 & 4180.37 \\
          & \multicolumn{1}{c|}{\textbf{DB}} & \textbf{4249} & \textbf{2.55} & \textbf{3304} & \textbf{3.52} & \textbf{544} & \textbf{0.39} & \textbf{508} & \textbf{0.44} & \textbf{615} & \textbf{0.40} & \textbf{11107} & \textbf{13.42} \\
    \hline
    \multirow{3}[2]{*}{\textbf{[1,2]}} & \multicolumn{1}{c|}{\textbf{CS}} & 864528 & 554.43 & 3416944 & 2100.07 & 42607 & 29.18 & 317550 & 207.11 & 132992 & 85.12 & 9637557 & 5899.29 \\
          & \multicolumn{1}{c|}{\textbf{DBN}} & 864528 & 477.71 & 3416944 & 1590.33 & 42607 & 20.86 & 317550 & 178.75 & 132992 & 62.52 & 9637557 & 4229.89 \\
          & \multicolumn{1}{c|}{\textbf{DB}} & \textbf{4291} & \textbf{2.63} & \textbf{3170} & \textbf{3.45} & \textbf{543} & \textbf{0.42} & \textbf{473} & \textbf{0.46} & \textbf{645} & \textbf{0.42} & \textbf{10715} & \textbf{13.24} \\
    \hline
    \multirow{3}[2]{*}{\textbf{[2,3]}} & \multicolumn{1}{c|}{\textbf{CS}} & 864528 & 564.79 & 3416944 & 2098.74 & 42607 & 28.19 & 317550 & 204.90 & 132992 & 84.15 & 9637557 & 5904.34 \\
          & \multicolumn{1}{c|}{\textbf{DBN}} & 864528 & 452.25 & 3416944 & 1502.59 & 42607 & 19.72 & 317550 & 170.06 & 132992 & 60.38 & 9637557 & 4076.98 \\
          & \multicolumn{1}{c|}{\textbf{DB}} & \textbf{3963} & \textbf{2.49} & \textbf{3162} & \textbf{3.45} & \textbf{566} & \textbf{0.41} & \textbf{428} & \textbf{0.40} & \textbf{646} & \textbf{0.41} & \textbf{9783} & \textbf{12.75} \\
    \hline
    \multirow{3}[2]{*}{\textbf{[1,1]}} & \multicolumn{1}{c|}{\textbf{CS}} & 864528 & 563.24 & 3416944 & 2097.62 & 42607 & 27.29 & 317550 & 204.26 & 132992 & 84.17 & 9637557 & 5877.44 \\
          & \multicolumn{1}{c|}{\textbf{DBN}} & 864528 & 518.04 & 3416944 & 1629.15 & 42607 & 19.47 & 317550 & 182.59 & 132992 & 64.14 & 9637557 & 4292.83 \\
          & \multicolumn{1}{c|}{\textbf{DB}} & \textbf{2320} & \textbf{1.61} & \textbf{3331} & \textbf{3.46} & \textbf{492} & \textbf{0.36} & \textbf{421} & \textbf{0.41} & \textbf{642} & \textbf{0.40} & \textbf{9804} & \textbf{12.58} \\
    \hline
    \multirow{3}[2]{*}{\textbf{[3,2]}} & \multicolumn{1}{c|}{\textbf{CS}} & 864528 & 555.61 & 3416944 & 2099.28 & 42607 & 27.00 & 317550 & 205.13 & 132992 & 84.52 & 9637557 & 5913.31 \\
          & \multicolumn{1}{c|}{\textbf{DBN}} & 864528 & 448.32 & 3416944 & 1502.94 & 42607 & 18.67 & 317550 & 171.57 & 132992 & 59.71 & 9637557 & 4077.99 \\
          & \multicolumn{1}{c|}{\textbf{DB}} & \textbf{1617} & \textbf{1.29} & \textbf{3397} & \textbf{3.56} & \textbf{531} & \textbf{0.39} & \textbf{410} & \textbf{0.39} & \textbf{590} & \textbf{0.38} & \textbf{9164} & \textbf{12.44} \\
    \hline
    \multirow{3}[2]{*}{\textbf{[2,1]}} & \multicolumn{1}{c|}{\textbf{CS}} & 864528 & 555.43 & 3416944 & 2102.39 & 42607 & 27.00 & 317550 & 204.46 & 132992 & 84.13 & 9637557 & 5912.66 \\
          & \multicolumn{1}{c|}{\textbf{DBN}} & 864528 & 483.04 & 3416944 & 1592.71 & 42607 & 18.98 & 317550 & 182.45 & 132992 & 62.60 & 9637557 & 4232.63 \\
          & \multicolumn{1}{c|}{\textbf{DB}} & \textbf{1590} & \textbf{1.30} & \textbf{3392} & \textbf{3.55} & \textbf{545} & \textbf{0.40} & \textbf{382} & \textbf{0.39} & \textbf{633} & \textbf{0.41} & \textbf{8896} & \textbf{12.32} \\
    \hline
    \multirow{3}[2]{*}{\textbf{[3,1]}} & \multicolumn{1}{c|}{\textbf{CS}} & 864528 & 558.94 & 3416944 & 2109.63 & 42607 & 27.05 & 317550 & 213.56 & 132992 & 86.65 & 9637557 & 5957.71 \\
          & \multicolumn{1}{c|}{\textbf{DBN}} & 864528 & 475.44 & 3416944 & 1555.26 & 42607 & 19.20 & 317550 & 184.76 & 132992 & 63.04 & 9637557 & 4179.39 \\
          & \multicolumn{1}{c|}{\textbf{DB}} & \textbf{1082} & \textbf{1.07} & \textbf{3469} & \textbf{3.59} & \textbf{472} & \textbf{0.36} & \textbf{432} & \textbf{0.45} & \textbf{600} & \textbf{0.40} & \textbf{8744} & \textbf{12.24} \\
    \hline
    \multirow{3}[2]{*}{\textbf{[5,1]}} & \multicolumn{1}{c|}{\textbf{CS}} & 864528 & 560.19 & 3416944 & 2110.52 & 42607 & 27.08 & 317550 & 206.32 & 132992 & 84.59 & 9637557 & 5966.70 \\
          & \multicolumn{1}{c|}{\textbf{DBN}} & 864528 & 469.49 & 3416944 & 1515.05 & 42607 & 18.81 & 317550 & 178.64 & 132992 & 60.76 & 9637557 & 4145.01 \\
          & \multicolumn{1}{c|}{\textbf{DB}} & \textbf{879} & \textbf{0.94} & \textbf{3569} & \textbf{3.65} & \textbf{479} & \textbf{0.38} & \textbf{392} & \textbf{0.38} & \textbf{508} & \textbf{0.35} & \textbf{8608} & \textbf{12.22} \\
    \hline
    \multirow{3}[2]{*}{\textbf{[7,1]}} & \multicolumn{1}{c|}{\textbf{CS}} & 864528 & 565.02 & 3416944 & 2119.74 & 42607 & 27.23 & 317550 & 208.69 & 132992 & 85.38 & 9637557 & 6004.51 \\
          & \multicolumn{1}{c|}{\textbf{DBN}} & 864528 & 461.46 & 3416944 & 1482.16 & 42607 & 18.74 & 317550 & 180.49 & 132992 & 59.88 & 9637557 & 4123.34 \\
          & \multicolumn{1}{c|}{\textbf{DB}} & \textbf{1080} & \textbf{1.20} & \textbf{3546} & \textbf{3.66} & \textbf{470} & \textbf{0.34} & \textbf{382} & \textbf{0.38} & \textbf{544} & \textbf{0.38} & \textbf{8643} & \textbf{12.29} \\
    \hline
    \multirow{3}[2]{*}{\textbf{[10,1]}} & \multicolumn{1}{c|}{\textbf{CS}} & 864528 & 572.12 & 3416944 & 2118.36 & 42607 & 26.94 & 317550 & 213.33 & 132992 & 85.57 & 9637557 & 6005.74 \\
          & \multicolumn{1}{c|}{\textbf{DBN}} & 864528 & 460.56 & 3416944 & 1461.84 & 42607 & 18.75 & 317550 & 176.00 & 132992 & 59.43 & 9637557 & 4126.60 \\
          & \multicolumn{1}{c|}{\textbf{DB}} & \textbf{1270} & \textbf{1.18} & \textbf{3622} & \textbf{3.73} & \textbf{435} & \textbf{0.33} & \textbf{459} & \textbf{0.45} & \textbf{547} & \textbf{0.38} & \textbf{9607} & \textbf{12.90} \\
    \hline
    \multirow{3}[2]{*}{\textbf{[25,1]}} & \multicolumn{1}{c|}{\textbf{CS}} & 864528 & 575.71 & 3416944 & 2121.00 & 42607 & 27.25 & 317550 & 214.90 & 132992 & 85.68 & 9637557 & 6025.70 \\
          & \multicolumn{1}{c|}{\textbf{DBN}} & 864528 & 457.59 & 3416944 & 1497.55 & 42607 & 19.11 & 317550 & 176.45 & 132992 & 60.98 & 9637557 & 4229.99 \\
          & \multicolumn{1}{c|}{\textbf{DB}} & \textbf{1523} & \textbf{1.35} & \textbf{3668} & \textbf{3.84} & \textbf{500} & \textbf{0.41} & \textbf{505} & \textbf{0.47} & \textbf{594} & \textbf{0.42} & \textbf{10683} & \textbf{13.67} \\
    \hline
    \multirow{3}[2]{*}{\textbf{[50,1]}} & \multicolumn{1}{c|}{\textbf{CS}} & 864528 & 578.59 & 3416944 & 2122.83 & 42607 & 27.13 & 317550 & 212.13 & 132992 & 85.56 & 9637557 & 6046.18 \\
          & \multicolumn{1}{c|}{\textbf{DBN}} & 864528 & 470.39 & 3416944 & 1527.57 & 42607 & 19.49 & 317550 & 171.24 & 132992 & 61.58 & 9637557 & 4321.96 \\
          & \multicolumn{1}{c|}{\textbf{DB}} & \textbf{1439} & \textbf{1.32} & \textbf{3662} & \textbf{3.86} & \textbf{486} & \textbf{0.37} & \textbf{500} & \textbf{0.51} & \textbf{599} & \textbf{0.42} & \textbf{11619} & \textbf{14.20} \\
    \hline
    \multirow{3}[2]{*}{\textbf{[100,1]}} & \multicolumn{1}{c|}{\textbf{CS}} & 864528 & 601.96 & 3416944 & 2126.83 & 42607 & 27.26 & 317550 & 215.27 & 132992 & 85.95 & 9637557 & 6060.87 \\
          & \multicolumn{1}{c|}{\textbf{DBN}} & 864528 & 500.53 & 3416944 & 1565.47 & 42607 & 19.84 & 317550 & 175.26 & 132992 & 63.33 & 9637557 & 4462.43 \\
          & \multicolumn{1}{c|}{\textbf{DB}} & \textbf{1375} & \textbf{1.24} & \textbf{3630} & \textbf{3.79} & \textbf{481} & \textbf{0.39} & \textbf{508} & \textbf{0.46} & \textbf{628} & \textbf{0.43} & \textbf{12667} & \textbf{14.57} \\
    \hline
    \multirow{3}[2]{*}{\textbf{Impr}} & \textbf{CS$\rightarrow$DBN} & \textbf{-} & \textbf{17.54\%} & \textbf{-} & \textbf{27.76\%} & \textbf{-} & \textbf{29.56\%} & \textbf{-} & \textbf{17.68\%} & \textbf{-} & \textbf{28.30\%} & \textbf{-} & \textbf{29.42\%} \\
    
    & \textbf{DBN$\rightarrow$DB} & \textbf{99.69\%} & \textbf{99.60\%} & \textbf{99.90\%} & \textbf{99.76\%} & \textbf{98.73\%} & \textbf{97.93\%} & \textbf{99.84\%} & \textbf{99.72\%} & \textbf{99.53\%} & \textbf{99.32\%} & \textbf{99.87\%} & \textbf{99.66\%} \\
          & \textbf{CS$\rightarrow$DB} & \textbf{99.69\%} & \textbf{99.67\%} & \textbf{99.90\%} & \textbf{99.83\%} & \textbf{98.73\%} & \textbf{98.54\%} & \textbf{99.84\%} & \textbf{99.78\%} & \textbf{99.53\%} & \textbf{99.51\%} & \textbf{99.87\%} & \textbf{99.76\%} \\
    \hline
    \end{tabular}}%
  \label{time_cmp}%
\end{table*}%

Results are shown in Table \ref{time_cmp}. As can be seen in the last row of this table, the polynomial model, even evaluating the same number of zeros (the searching method is the Zeroin algorithm \cite{Dekker69,Brent73}) gets an average of 25\% training time saving respect to the hyperbolic model in Cost-Sensitive Boosting. On the other hand, the Conditional Search method achieves a reduction over 99.5\% on the total number of evaluated zeros, driving the full version of AdaBoostDB to consume only 0.49\% of the time on average used by Cost-Sensitive Boosting. That is, it is more than 200 times faster.

\subsection{Real-world dataset}
\label{sub_sec:real_world}

As last experiment we have trained, with AdaBoostDB as learning algorithm, a simple mono-stage face detector using Haar-like features \cite{ViolaJones04}, a kind of real-world asymmetric problem in which boosting is commonly used. For this purpose we have used a balanced subset (i.e. with the same number of positive and negative samples) from the CBCL training face and non-face datasets \cite{Heisele00,Alvira01}. Obtained results are shown in Table \ref{CBCL_table} confirming, once again, the consistent cost-sensitive behavior of the classifiers trained with AdaBoostDB in different scenarios.

\begin{table*}[htbp]
\footnotesize
    \centering
  \caption{AdaBoostDB asymmetric behavior (false negatives, false positives, classification error and normalized expected cost) for each cost combination over the CBCL example dataset.}    
        \scalebox{0.9}{
  \begin{tabular}{|c|c|c|c|c|}
    \hline
    \multirow{2}[4]{*}{\textbf{Cost}} & \multicolumn{4}{c|}{\textbf{CBCL}} \\
\cline{2-5} & \multicolumn{1}{c|}{\textbf{FN}} & \multicolumn{1}{c|}{\textbf{FP}} & \multicolumn{1}{c|}{\textbf{CE}} & \multicolumn{1}{c|}{\textbf{NEC}} \\
    \hline
    \textbf{[1,100]} & 4.40$\cdot10^{-1}$ & 7.50$\cdot10^{-3}$ & 2.24$\cdot10^{-1}$ & 1.18$\cdot10^{-2}$ \\
    \textbf{[1,50]} & 2.81$\cdot10^{-1}$ & 1.25$\cdot10^{-2}$ & 1.47$\cdot10^{-1}$ & 1.78$\cdot10^{-2}$ \\
    \textbf{[1,25]} & 2.35$\cdot10^{-1}$ & 1.08$\cdot10^{-2}$ & 1.23$\cdot10^{-1}$ & 1.95$\cdot10^{-2}$ \\
    \textbf{[1,10]} & 2.78$\cdot10^{-1}$ & 1.00$\cdot10^{-2}$ & 1.44$\cdot10^{-1}$ & 3.43$\cdot10^{-2}$ \\
    \textbf{[1,7]} & 1.72$\cdot10^{-1}$ & 1.58$\cdot10^{-2}$ & 9.38$\cdot10^{-2}$ & 3.53$\cdot10^{-2}$ \\
    \textbf{[1,5]} & 1.20$\cdot10^{-1}$ & 1.00$\cdot10^{-2}$ & 6.50$\cdot10^{-2}$ & 2.83$\cdot10^{-2}$ \\
    \textbf{[1,3]} & 1.17$\cdot10^{-1}$ & 2.33$\cdot10^{-2}$ & 7.00$\cdot10^{-2}$ & 4.67$\cdot10^{-2}$ \\
    \textbf{[1,2]} & 9.92$\cdot10^{-2}$ & 2.83$\cdot10^{-2}$ & 6.38$\cdot10^{-2}$ & 5.19$\cdot10^{-2}$ \\
    \textbf{[2,3]} & 7.08$\cdot10^{-2}$ & 1.83$\cdot10^{-2}$ & 4.46$\cdot10^{-2}$ & 3.93$\cdot10^{-2}$ \\
    \textbf{[1,1]} & 8.50$\cdot10^{-2}$ & 2.25$\cdot10^{-2}$ & 5.38$\cdot10^{-2}$ & 5.38$\cdot10^{-2}$ \\
    \textbf{[3,2]} & 6.92$\cdot10^{-2}$ & 4.25$\cdot10^{-2}$ & 5.58$\cdot10^{-2}$ & 5.85$\cdot10^{-2}$ \\
    \textbf{[2,1]} & 8.17$\cdot10^{-2}$ & 2.58$\cdot10^{-2}$ & 5.38$\cdot10^{-2}$ & 6.31$\cdot10^{-2}$ \\
    \textbf{[3,1]} & 4.25$\cdot10^{-2}$ & 3.42$\cdot10^{-2}$ & 3.83$\cdot10^{-2}$ & 4.04$\cdot10^{-2}$ \\
    \textbf{[5,1]} & 9.08$\cdot10^{-2}$ & 2.92$\cdot10^{-2}$ & 6.00$\cdot10^{-2}$ & 8.06$\cdot10^{-2}$ \\
    \textbf{[7,1]} & 4.17$\cdot10^{-2}$ & 6.58$\cdot10^{-2}$ & 5.38$\cdot10^{-2}$ & 4.47$\cdot10^{-2}$ \\
    \textbf{[10,1]} & 4.25$\cdot10^{-2}$ & 5.00$\cdot10^{-2}$ & 4.63$\cdot10^{-2}$ & 4.32$\cdot10^{-2}$ \\
    \textbf{[25,1]} & 3.33$\cdot10^{-2}$ & 7.17$\cdot10^{-2}$ & 5.25$\cdot10^{-2}$ & 3.48$\cdot10^{-2}$ \\
    \textbf{[50,1]} & 2.08$\cdot10^{-2}$ & 1.29$\cdot10^{-1}$ & 7.50$\cdot10^{-2}$ & 2.30$\cdot10^{-2}$ \\
    \textbf{[100,1]} & 3.92$\cdot10^{-2}$ & 1.65$\cdot10^{-1}$ & 1.02$\cdot10^{-1}$ & 4.04$\cdot10^{-2}$ \\
    \hline
    \end{tabular}}%
  \label{CBCL_table}%
\end{table*}%

\section{Conclusions}
\label{sec:Conclusions}

In this paper we have presented, derived and empirically tested a new cost-sensitive AdaBoost scheme, AdaBoostDB, based on double-base exponential error bounds. Sharing an equivalent theoretical root with Cost-Sensitive Boosting \cite{MasnadiVasconcelos11} and opposed to the most of other asymmetric approaches in the literature, AdaBoostDB is supported by a full theoretical derivation that makes it possible preserve all the formal guarantees of the original AdaBoost for a general asymmetric scenario. 

Our approach is based on three basic mainstays: the double-base perspective, a derivation scheme based on the generalized boosting framework \cite{SchapireSinger99} (instead of the Statistical View of Boosting used in \cite{MasnadiVasconcelos11}) and a polynomial model for the problem (opposed to the hyperbolic one proposed in \cite{MasnadiVasconcelos11}). These distinctive features, as a whole, also enable a Conditional Search method to increase compactness, ease and efficiency of the algorithm. As a consequence, AdaBoostDB training consumes only 0.49\% of the time on average needed by Cost-Sensitive AdaBoost to reach the same solution. This computational advantage (200 times faster) can make a difference in applications coping with a huge number (hundreds of thousands, even millions) of weak hypothesis, as object detection in computer vision.

From this point, next steps of our research will require a thorough comparison between AdaBoostDB/Cost-Sensitive AdaBoost and AdaBoost with asymmetric weight initialization \cite{LandesaAlba12} (the other fully-theoretical asymmetric boosting model in the literature) in order to clarify, both theoretically and practically, the different properties, advantages and disadvantages of each asymmetry model.

\vskip 0.2in

\section{Acknowledgements}
\label{sec:acknowledgements}{This work was partially supported by the European Regional Development Fund (ERDF) and the Galician Regional Government under project CN 2012/260
``Consolidation of Research Units: AtlantTIC'', and by the Galician Regional Funds through the Research contract CN2011/019 (Modalidade: Grupos de Referencia Competitiva 2007).}

\bibliographystyle{model1-num-names}
\bibliography{AdaBoostDB}

\end{document}